*Article*

# Cascade Decoders-Based Autoencoders for Image Reconstruction

Honggui Li [1,*], Maria Trocan [2], Mohamad Sawan [3,4] and Dimitri Galayko [5]

[1]Yangzhou University; hgli@yzu.edu.cn
[2]Institut Supérieur d'Électronique de Paris; maria.trocan@isep.fr
[3]Polytechnique Montreal; mohamad.sawan@polymtl.ca
[4]Westlake University; sawan@westlake.edu.cn
[5]Sorbonne University; dimitri.galayko@sorbonne-universite.fr
**\*** Correspondence: hgli@yzu.edu.cn; Tel.: +86-18952781178

**Abstract:** Autoencoders are composed of coding and decoding units, hence they hold the inherent potential of high-performance data compression and signal compressed sensing. The main disadvantages of current autoencoders comprise the following several aspects: the research objective is not lossless data reconstruction but efficient feature representation; the performance evaluation of data recovery is neglected; it is hard to achieve lossless data reconstruction by pure autoencoders, even by pure deep learning. This paper aims at image reconstruction of autoencoders, employs cascade decoders-based autoencoders, perfects the performance of image reconstruction, approaches gradually lossless image recovery, and provides solid theory and application basis for autoencoders-based image compression and compressed sensing. The proposed serial decoders-based autoencoders include the architectures of multi-level decoders and the related optimization algorithms. The cascade decoders consist of general decoders, residual decoders, adversarial decoders and their combinations. The effectiveness of residual cascade decoders for image reconstruction is proven in mathematics. It is evaluated by the experimental results that the proposed autoencoders outperform the classical autoencoders in the performance of image reconstruction.

## 1. Introduction

Because deep learning highly efficiently achieves the rules and features from input data by multi-layer stacked neural networks, it gains unprecedented-successful research and applications in the domain of data classification, recognition, compression and processing [1-2]. Although the theoretical research and engineering applications of deep learning develop maturely, it still has much space to improve and it does not attain the requirement of general artificial intelligence [1-2]. Hence, it is high time for researchers to utilize deep learning to upgrade the performance of data compression and signal compressed sensing.

Data reconstruction is the foundation of data compression and signal compressed sensing. It contains multifarious meanings by understanding from broad sense. In this paper, it denotes that high-dimensional original data are firstly mapped into a low-dimensional space and then they are recovered. Although the classical methods of data compression and signal compressed sensing are full-blown, it is still necessary to investigate the new algorithms based on deep learning. Currently, merely some of the components of traditional data compression methods, such as prediction coding, transformation coding and quantization coding, are replaced by deep learning methods. The principal difficulty is that lossless data reconstruction of pure deep learning-based methods is currently not attained. In consideration of the powerful capability of deep learning, this article will explore the new approaches to data reconstruction via pure deep learning-based methods.

Autoencoders (AE) are a classical architecture of deep neural networks, which firstly project high-dimensional data into a low-dimensional latent space according to a given rule and then reconstruct the original data from latent space by minimizing the reconstruction error [3-6]. Autoencoders possess many theoretical models, such as sparse autoencoders, convolutional autoencoders, variational autoencoders (VAE), adversarial autoencoders (AAE), Wasserstein autoencoders (WAE), graphical autoencoders, extreme learning autoencoders, integral learning autoencoders, inverse function autoencoders, recursive or recurrent autoencoders, double or couple autoencoders, denoising autoencoders, generative autoencoders, fuzzy autoencoders, nonnegative autoencoders, binary autoencoders, quantum autoencoders, linear autoencoders, blind autoencoders, group autoencoders, kernel autoencoders and etcetera [3-6]. Autoencoders attain extensive research and applications in the domain of classification, recognition, encoding, sensing and processing [3-6]. Because autoencoders comprise encoding and decoding units, they hold the potential of high-performance data compression and signal compressed sensing [7-8]. Classical autoencoders can be called as narrow autoencoders. Other deep learning-based methods of data compression and signal compressed sensing can be named as generalized autoencoders, because they contain encoding and decoding components and each component can introduce autoencoders unit [9-10]. Narrow and generalized autoencoders-based approaches of data compression and signal compressed sensing can provide better performance of data reconstruction than the classical approaches [7-10].

However, the current research of autoencoders exists the following problems: research target is not lossless data reconstruction but efficient feature representation; the independent evaluation of the performance of data reconstruction is neglected; the performance of data reconstruction needs to be improved; it is hard to attain lossless data reconstruction [11-12]. For instance, the performance of data reconstruction of AAE, one of the most advanced autoencoders, is shown in Fig. 1 [13]. The horizontal axis is the dimension of the latent space and the vertical axis is the average structural similarity (SSIM) of reconstructed images in comparison with original images. It is indicated in Fig. 1 that the performance of data reconstruction of AAE increases while the dimension of hidden space increases and it is difficult to reach lossless data reconstruction. Actually and currently, pure deep learning-based methods of data compression and signal compressed sensing cannot attain lossless data reconstruction.

This manuscript attempts to regard lossless data reconstruction as research goal of autoencoders, independently assesses the performance of data reconstruction of autoencoders, enhances the quality of data reconstruction of autoencoders, gradually approaches lossless data reconstruction of autoencoders, and builds solid theory and application foundation of data compression and signal compressed sensing of autoencoders.

This article proposes serial decoders-based autoencoders for image reconstruction. The main contribution of this paper is the introduction of cascade decoders into autoencoders, including theoretical architectures, optimization problems and training algorithms. The components of serial decoders consist of general decoders, residual decoders, adversarial decoders and their combinations. The effectiveness of residual serial decoders for image reconstruction is proven in mathematics. Because AAE, VAE and WAE are the state-of-the-art autoencoders, this article focuses on their cascade decoders-based   versions.
The rest part of this article is organized as follows. The related work is summarized in section 2, the theoretical foundations are established in section 3, the simulation experiments are designed in section 4, and the final conclusions are drawn in section 5.

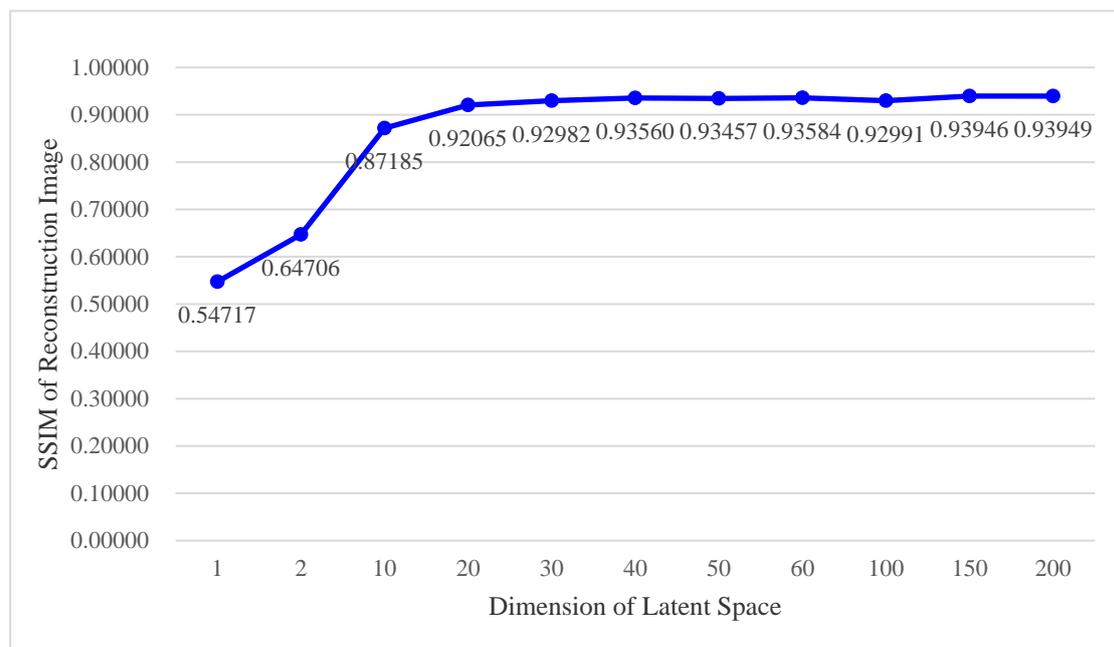

**Figure 1.** The performance of data reconstruction of AAE.

## 2. Related-work

Narrow autoencoders-based data compression and signal compressing progress rapidly [7-8, 14-19]. Firstly, autoencoders have been studied and applied in the compression of medical signal, navigation data and quantum states [7, 14-16]. For example, Wu Tong et al. propose autoencoders based compression method of brain neural signal [7]. Yildirim Ozal et al. utilize convolutional autoencoders to compress electrocardiosignal [14]. Lokukaluge P. Perera et al. employ linear autoencoders to compress navigation data [15]. Romero Jonathan et al. use quantum autoencoders to compress quantum states [16]. Secondly, autoencoders have already been studied and applied in the compressed sensing of biomedical signal, image and sensor data [8, 17-19]. For instance, Gogna Anupriya et al. utilize stacked and label-consistent autoencoders to reconstruct electrocardiosignal and electroencephalogram [8]. Biao Sun et al. use binary autoencoders for compressed sensing of neural signal [17]. Majumdar Angshul wields autoencoders to reconstruct magnetic resonant image [18]. Han Tao et al. adopt sparse autoencoders to reconstruct sensor signal [19].

The important development of narrow autoencoders also includes: Wasserstein autoencoder, inverse function autoencoder, and graphical autoencoders [20-22]. For example, Ilya Tolstikhin et al. raise Wasserstein autoencoders, which are generalized adversarial autoencoders, utilize Wasserstein distance to measure the difference between data model distribution and target distribution, and gain better performance of data reconstruction than classical variational autoencoders and adversarial autoencoders [20]. Yimin Yang et al. employ inverse activation function and pseudo inverse matrix to get the analysis representation of the network parameters of autoencoders for dimension reduction and reconstruction of image data, and improve the data reconstruction performance of autoencoders [21]. Majumdar Angshul presents graphical autoencoders, uses graphical regularization for data denoising, clustering and classification, and attains better data reconstruction performance than classical autoencoders [22].

Generalized autoencoders-based data compression and signal compressed sensing also achieve significant evolution [9-10, 23]. These methods usually utilize multi-level autoencoders to overcome the disadvantage that the single-level autoencoders cannot reach lossless data reconstruction, or use autoencoders to replace one unit of the classical data compression and signal compressed sensing, such as the prediction, transformation or quantization unit of data compression and the measurement or recovery unit of signal compressed sensing. For instance, George Toderici et al. wield two-level autoencoders for image compression. The first-level autoencoders compress image blocks and the second-level autoencoders compress the recovery residuals of the first-level autoencoders. It makes up the disadvantage that single-level autoencoders cannot implement lossless data reconstruction to a great degree [9]. Oren Rippel et al. adopt multi-level autoencoders to implement the transformation coding unit of video compression. The first-level autoencoders compress the prediction residuals and the next-level autoencoders respectively compress the reconstruction residuals of the previous-level autoencoders to a great extent [10]. Majid Sepahvand et al. employ autoencoders to implement the prediction coding unit of compressed sensing of sensor signal [23].

The main research advances in generalized autoencoders also comprise using other architectures of deep neural networks to implement data compression and signal compressed sensing [24-27]. In data compression, these methods usually wield deep neural networks to substitute the prediction, transformation or quantization unit of classical methods. In signal compressed sensing, these methods usually wield deep neural networks to substitute the measurement or recovery unit of classical methods. For example, Jiahao Li et al. utilize fully-connected deep neural networks to realize intra prediction coding unit of video compression [25]. Guo Lu et al. adopt deep convolutional neural networks to replace the transformation coding unit of video compression [26]. Wenxue Cui et al. employ deep convolutional neural network to accomplish the sampling and reconstruction units of image compressed sensing [27]. This paper focuses on narrow autoencoders, incorporates multi-level decoders into autoencoders, and boosts the performance of data reconstruction. To the best of our knowledge, cascade decoders in autoencoders have never been studied. Although serial autoencoders have already been investigated, serial decoders in autoencoders play a more important role in data reconstruction. In addition, Tero Karras et al. progressively train generative adversarial networks by gradually increasing the layer number of generator and discriminator to improve the quality, stability and variability [28]. This method will be borrowed for progressively training the proposed cascade decoders-based autoencoders.

## 3. Theory

### 3.1. Notations and Abbreviations

For the convenience of content description, parts of the mathematical notations and abbreviations adopted in this manuscript are listed in the following Tab. 1.

**Table 1.** Mathematical notations and abbreviations.

| Notations and Abbreviations | Meanings |
|---|---|
| AE | autoencoders |
| AAE/VAE/WAE | adversarial/variational/Wasserstein AE |
| CD | cascade decoders |
| GCD/RCD/ACD/RACD | general/residual/adversarial/residual-adversarial CD |
| CD/GCDAE/RCDAE/ACDAE/RACDAE | CD/GCD/RCD/ACD/RACD-based AE |
| CDAAE/GCDAAE/RCDAAE/ACDAAE/RACDAAE | CD/GCD/RCD/ACD/RACD-based AAE |
| CDVAE/GCDVAE/RCDVAE/ACDVAE/RACDVAE | CD/GCD/RCD/ACD/RACD-based VAE |
| CDWAE/GCDWAE/RCDWAE/ACDWAE/RACDWAE | CD/GCD/RCD/ACD/RACD-based WAE |
| E/D/DC | encoder/decoder/discriminator |
| **x/y/z** | original/reconstructed/latent sample |

*3.2. Recall of classical autoencoders*

The architecture of the classical autoencoders is illustrated in Fig. 2. The classical autoencoders are composed of two units: Encoder and Decoder. Encoder reduces the high-dimensional input data to a low-dimensional representation and Decoder reconstructs the high-dimensional data from the low-dimensional representation. The classical autoencoder can be described by the following formulas:

$$\begin{aligned} \mathbf{z} &= \mathrm{E}(\mathbf{x}) \\ \mathbf{y} &= \mathrm{D}(\mathbf{z}) \\ \mathbf{x}, \mathbf{y} &\in R^H; \mathbf{z} \in R^L \\ H &\gg L \end{aligned} \qquad (1)$$

where:
$\mathbf{x}$ is the high-dimensional input data; Taking image data as an example, $\mathbf{x}$ is the normalized version of original image for the convenience of numerical computation; Each element of original image is an integer in range [0, 255]; Each element of $\mathbf{x}$ is a real number in range [0, 1] or [-1, +1]; $\mathbf{x}$ with elements in range [0, 1] can be understood as probability variable; $\mathbf{x}$ can also be regarded as a vector which is a reshaping version of an image matrix;
$\mathbf{z}$ is the low-dimensional representation in a latent space;
$\mathbf{y}$ is the high-dimensional data, such as a reconstruction image;
E is the Encoder;
D is the Decoder;
$H$ is the dimension of $\mathbf{x}$ or $\mathbf{y}$; For image data, $H$ is equal to the product of image width and height;
$L$ is the dimension of $\mathbf{z}$ and $L$ is far less than $H$.

The classical autoencoders can be resolved by the following optimization problem:

$$(\boldsymbol{\theta}, \mathbf{z}, \mathbf{y}) = \underset{\boldsymbol{\theta}, \mathbf{y}, \mathbf{z}}{\arg\min} \|\mathbf{y} - \mathbf{x}\|_2^2$$
$$\text{s.t. } \mathbf{z} = \mathrm{E}(\mathbf{x}), \mathbf{y} = \mathrm{D}(\mathbf{z}), C_z = \|\mathbf{z} - \mathbf{z}_g\|_2^2 < \delta_z, C_y = \|\mathbf{y} - \mathbf{D}_y \mathbf{s}_y\|_2^2 + \lambda_y \|\mathbf{s}_y\|_1 < \delta_y \qquad (2)$$

where:
$\boldsymbol{\theta}$ is the parameters of autoencoders including the parameters of Encoder and Decoder;
$C_z$ is the constraint on low-dimensional representation $\mathbf{z}$; for example, $\mathbf{z}$ satisfies a given probability distribution; it has been considered to match a known distribution by the classical adversarial autoencoders, variational autoencoders and Wasserstein autoencoders;
$\mathbf{z}_g$ is a related variable which meets a given distribution;
$\delta_z$ is a small constant;
$C_y$ is the constraint on high-dimensional reconstruction data $\mathbf{y}$; for instance, $\mathbf{y}$ meets a prior of local smoothness or nonlocal similarity; autoencoders require y to reconstruct $\mathbf{x}$ based on the prior to a great extent; other constraints, such as sparsity and low-rank properties of high-dimensional reconstruction data, can also be utilized; hereby, sparse prior is taken as an example;
$\mathbf{D}_y$ is a matrix of sparse dictionary;
$\mathbf{s}_y$ is a vector of sparse coefficients;
$\lambda_y$ is a small constant;

$\delta_y$ is a small constant.

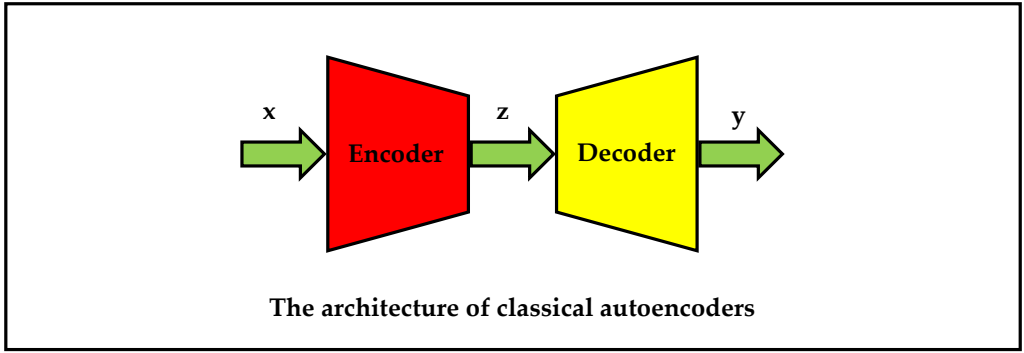

**Figure 2.** The architecture of classical autoencoders.

*3.3. Proposed cascade decoders based autoencoders*

The framework of the proposed cascade decoders-based autoencoders (CDAE) is exhibited in Fig. 3. The framework consists of two components: Encoder and cascade decoders. Encoder is similar to that in the classical autoencoder. Cascade decoders comprise N serial decoders from decoder 1 to N. The framework can be depicted by the following expressions:

$$\mathbf{z} = E(\mathbf{x})$$
$$\mathbf{y}_n = \begin{cases} D_1(\mathbf{z}), n = 1 \\ D_n(\mathbf{y}_{n-1}), n = 2, ..., N \end{cases}, \mathbf{y} = \mathbf{y}_N, \qquad (3)$$

where:

$D_n$ is the n-th Decoder;

$\mathbf{y}_n$ is the reconstruction data of $D_n$.

The reconstruction data can be solved by the following optimization problem:

$$(\boldsymbol{\theta}; \mathbf{z}; \mathbf{y}_1, \cdots, \mathbf{y}_N; \mathbf{y}) = \underset{\boldsymbol{\theta}; \mathbf{z}; \mathbf{y}_1, \cdots, \mathbf{y}_N; \mathbf{y}}{\arg\min} \sum_{n=1}^{N} \|\mathbf{y}_n - \mathbf{x}\|_2^2$$
$$\text{s.t. } \mathbf{z} = E(\mathbf{x}); \mathbf{y}_1 = D_1(\mathbf{z}); \mathbf{y}_n = D_n(\mathbf{y}_{n-1}), n = 2, ..., N; , \qquad (4)$$
$$C_z; C_{y_n}, n = 1, ..., N; \mathbf{y} = \mathbf{y}_N$$

where:

$\boldsymbol{\theta}$ is the parameters of cascade decoders-based autoencoders;

$C_{y_n}$ is the constraint on $\mathbf{y}_n$.

For the purpose of gradually and serially training cascade decoders-based autoencoders, the optimization problem in Eq. 4 can be divided into the following sub optimization problems:

$$(\boldsymbol{\theta}_1; \mathbf{z}; \mathbf{y}_1) = \underset{\boldsymbol{\theta}_1; \mathbf{z}; \mathbf{y}_1}{\arg\min} \|\mathbf{y}_1 - \mathbf{x}\|_2^2$$
$$(\boldsymbol{\theta}_2; \mathbf{y}_2) = \underset{\boldsymbol{\theta}_2; \mathbf{y}_2}{\arg\min} \|\mathbf{y}_2 - \mathbf{x}\|_2^2$$
$$\cdots\cdots \qquad , \qquad (5)$$
$$(\boldsymbol{\theta}_N; \mathbf{y}_N; \mathbf{y}) = \underset{\boldsymbol{\theta}_N; \mathbf{y}_N; \mathbf{y}}{\arg\min} \|\mathbf{y}_N - \mathbf{x}\|_2^2$$
$$\text{s.t. } \mathbf{y} = E(\mathbf{x}); \mathbf{y}_1 = D_1(\mathbf{z}); \mathbf{y}_n = D_n(\mathbf{y}_{n-1}), n = 2, ..., N;$$
$$C_z; C_{y_n}, n = 1, ..., N; \mathbf{y} = \mathbf{y}_N$$

where:

$\boldsymbol{\theta}_1$ is the parameters of the Encoder and Decoder 1;

$\theta_2, \ldots$, and $\theta_N$ are the parameters of the Decoder 2 to N.

The proposed cascade decoders include general cascade decoders, residual cascade decoders, adversarial cascade decoders and their combinations. The general cascade decoders based autoencoders (GCDAE) have already been introduced in Fig. 3. The other cascade decoders are respectively elaborated as follows.

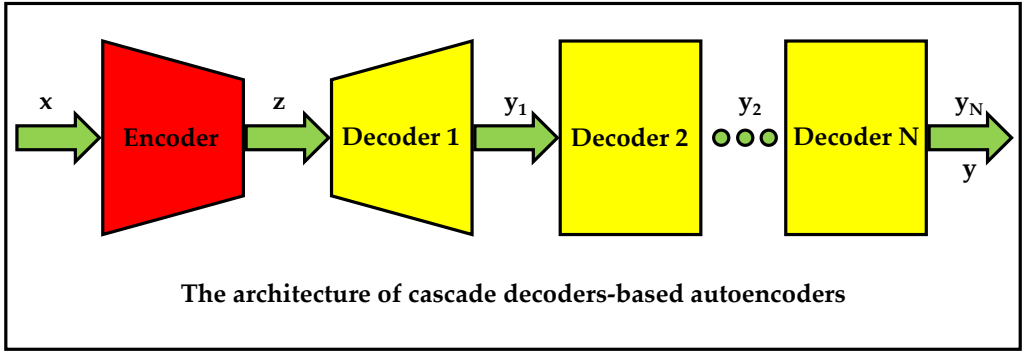

**Figure 3.** The architecture of cascade decoders-based autoencoders.

3.3.1. Residual cascade decoders-based autoencoders

The infrastructure of residual cascade decoders-based autoencoders (RCDAE) is demonstrated in Fig. 4. The blue signal flow is for the training phase, and the green signal flow is for the both phases of training and testing. Each decoder is a residual module. This architecture is different from the traditional residual network (ResNet) because the former has an extra training channel for residual computation.

The reconstruction data can be resolved by the following optimization problem:

$$\begin{aligned}(\theta; \mathbf{z}; \mathbf{r}_1, \cdots, \mathbf{r}_N; \mathbf{y}_1, \cdots, \mathbf{y}_N; \mathbf{y}) &= \underset{\theta; \mathbf{z}; \mathbf{y}_1, \cdots, \mathbf{y}_N; \mathbf{y}}{\arg\min} \sum_{n=1}^{N} \|\mathbf{x} - \mathbf{y}_{n-1} - \mathbf{r}_n\|_2^2 \\ \text{s.t. } &\mathbf{z} = E(\mathbf{x}); \mathbf{r}_1 = D_1(\mathbf{z}); \mathbf{r}_n = D_n(\mathbf{y}_{n-1}), n = 2, \ldots, N; \\ &C_{\mathbf{z}}; C_{\mathbf{y}_n}, n = 1, \ldots, N; \mathbf{y}_0 = \mathbf{0}; \mathbf{y}_n = \mathbf{r}_n + \mathbf{y}_{n-1}, n = 1, \ldots, N; \mathbf{y} = \mathbf{y}_N\end{aligned} \quad (6)$$

where:

$\mathbf{r}_n$ is the residual sample between $\mathbf{x}$ and $\mathbf{y}_n$;

$\mathbf{y}_0$ is the zero sample;

$\mathbf{y}$ is the final reconstruction sample.

For the purpose of gradually and serially training residual cascade decoders-based autoencoders, the optimization problem in Eq. 6 can be partitioned into the following sub optimization problems:

$$\begin{aligned}(\theta_1; \mathbf{z}; \mathbf{r}_1; \mathbf{y}_1) &= \underset{\theta_1; \mathbf{z}; \mathbf{r}_1; \mathbf{y}_1}{\arg\min} \|\mathbf{x} - \mathbf{y}_0 - \mathbf{r}_1\|_2^2 \\ (\theta_2; \mathbf{r}_2; \mathbf{y}_2) &= \underset{\theta_2; \mathbf{r}_2; \mathbf{y}_2}{\arg\min} \|\mathbf{x} - \mathbf{y}_1 - \mathbf{r}_2\|_2^2 \\ &\cdots\cdots \\ (\theta_N; \mathbf{r}_N; \mathbf{y}_N; \mathbf{y}) &= \underset{\theta_N; \mathbf{r}_N; \mathbf{y}_N; \mathbf{y}}{\arg\min} \|\mathbf{x} - \mathbf{y}_{N-1} - \mathbf{r}_N\|_2^2 \\ \text{s.t. } &\mathbf{z} = E(\mathbf{x}); \mathbf{r}_1 = D_1(\mathbf{z}); \mathbf{r}_n = D_n(\mathbf{y}_{n-1}), n = 2, \ldots, N; \\ &C_{\mathbf{z}}; C_{\mathbf{y}_n}, n = 1, \ldots, N; \mathbf{y}_0 = \mathbf{0}; \mathbf{y}_n = \mathbf{r}_n + \mathbf{y}_{n-1}, n = 1, \ldots, N; \mathbf{y} = \mathbf{y}_N\end{aligned} \quad (7)$$

The effectiveness of residual cascade decodes for image reconstruction can be proven as follows:

$$\begin{aligned}
&\mathbf{y}_n = \mathbf{r}_n + \mathbf{y}_{n-1}, n = 1,\cdots, N \\
&\Rightarrow \mathbf{x} - \mathbf{y}_n = (\mathbf{x} - \mathbf{y}_{n-1}) - \mathbf{r}_n \\
&\mathbf{r}_n \to (\mathbf{x} - \mathbf{y}_{n-1}) \Rightarrow \mu = \lim_{\mathbf{r}_n \to (\mathbf{x}-\mathbf{y}_{n-1})} \frac{\mathbf{r}_n}{(\mathbf{x}-\mathbf{y}_{n-1})} \to 1 \\
&\mathbf{r}_n \to (\mathbf{x}-\mathbf{y}_{n-1}) \overset{\mu \to 1, \varepsilon \to 0}{\Rightarrow} \mathbf{r}_n = \mu(\mathbf{x}-\mathbf{y}_{n-1}) + \varepsilon, \\
&\Rightarrow \mathbf{x}-\mathbf{y}_n = (\mathbf{x}-\mathbf{y}_{n-1}) - \mu(\mathbf{x}-\mathbf{y}_{n-1}) - \varepsilon = (1-\mu)(\mathbf{x}-\mathbf{y}_{n-1}) - \varepsilon \\
&\Rightarrow \|\mathbf{x}-\mathbf{y}_n\|_2 = \|(1-\mu)(\mathbf{x}-\mathbf{y}_{n-1}) - \varepsilon\|_2 \\
&\Rightarrow \|\mathbf{x}-\mathbf{y}_n\|_2 < \|(1-\mu)(\mathbf{x}-\mathbf{y}_{n-1})\|_2 + \|\varepsilon\|_2 \\
&\Rightarrow \|\mathbf{x}-\mathbf{y}_n\|_2 \overset{\varepsilon \to 0}{<} \|(1-\mu)(\mathbf{x}-\mathbf{y}_{n-1})\|_2 = |1-\mu|\|(\mathbf{x}-\mathbf{y}_{n-1})\|_2 \\
&\Rightarrow \|\mathbf{x}-\mathbf{y}_n\|_2 \overset{\mu \to 1}{<} 1 \cdot \|(\mathbf{x}-\mathbf{y}_{n-1})\|_2 = \|(\mathbf{x}-\mathbf{y}_{n-1})\|_2 \\
&\Rightarrow \|\mathbf{x}-\mathbf{y}_N\|_2 < \|\mathbf{x}-\mathbf{y}_{N-1}\|_2 < \cdots < \|\mathbf{x}-\mathbf{y}_2\|_2 < \|\mathbf{x}-\mathbf{y}_1\|_2
\end{aligned} \quad (8)$$

where $\mathbf{r}_n$ is close to $(\mathbf{x}-\mathbf{y}_{n-1})$ in the training phase in Eq. 6 and Fig. 4; $\mathbf{r}_n$ is the summation of a scaled $(\mathbf{x}-\mathbf{y}_{n-1})$ and a small error; $u$ is a scale coefficient which is approximate to 1; $\varepsilon$ is a error vector which is approximate to 0; reconstruction error decreases when the total number of decoders increases.

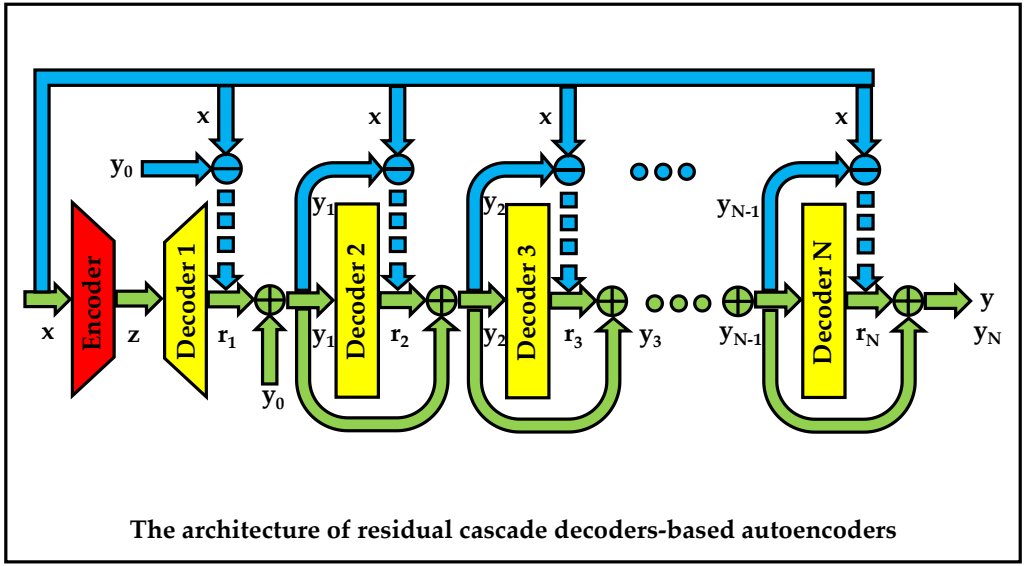

**The architecture of residual cascade decoders-based autoencoders**

**Figure 4.** The architecture of residual cascade decoders-based autoencoders.

3.3.2. Adversarial cascade decoders-based autoencoders

The architecture of adversarial cascade decoders-based autoencoders (ACDAE) is displayed in Fig. 5. The blue flow line represents the training phase, and the green flow represents the both phases of training and testing. Each Decoder is an adversarial module.

The reconstruction data can be solved by the following optimization problem:

$$\begin{aligned}
(\theta; \mathbf{z}; \mathbf{y}_1, \cdots, \mathbf{y}_N; \mathbf{y}) = \arg \min_{E;D} \max_{DC_1,\cdots,DC_N} \sum_{n=1}^{N} \left( \alpha_n M\left(\ln\left(DC_n(\mathbf{x})\right)\right) + \beta_n M\left(\ln\left(1 - DC_n(\mathbf{y}_n)\right)\right)\right) \\
\text{s.t. } \mathbf{z} = E(\mathbf{x}); \mathbf{y}_1 = D_1(\mathbf{z}); \mathbf{y}_n = D_n(\mathbf{y}_{n-1}), n = 2,\ldots,N; \\
C_z; \sum_{n=1}^{N} \|\mathbf{y}_n - \mathbf{x}\|_2^2 < \varepsilon, n = 1,\ldots,N; \\
\mathbf{y} = \mathbf{y}_N
\end{aligned} \quad (9)$$

where:

DC$_n$ is the n-th Discriminator;

$\alpha_n$ is a constant;

$\beta_n$ is a constant;

$\varepsilon$ is a small positive constant;

M is the mean operator.

For the sake of gradually and serially training adversarial cascade decoders-based autoencoders, the optimization problem in Eq. 9 can be divided into the following sub optimization problems:

$$(\boldsymbol{\theta}_1; \mathbf{z}; \mathbf{y}_1) = \arg\min_{E;D_1} \max_{DC_1} \left( \alpha_1 M\left(\ln\left(DC_1(\mathbf{x})\right)\right) + \beta_1 M\left(\ln\left(1 - DC_1(\mathbf{y}_1)\right)\right) \right)$$

$$(\boldsymbol{\theta}_2; \mathbf{y}_2) = \arg\min_{D_2} \max_{DC_2} \left( \alpha_2 M\left(\ln\left(DC_2(\mathbf{x})\right)\right) + \beta_2 M\left(\ln\left(1 - DC_2(\mathbf{y}_2)\right)\right) \right)$$

$$......$$

$$(\boldsymbol{\theta}_N; \mathbf{y}_N; \mathbf{y}) = \arg\min_{D_N} \max_{DC_N} \left( \alpha_N M\left(\ln\left(DC_N(\mathbf{x})\right)\right) + \beta_N M\left(\ln\left(1 - DC_N(\mathbf{y}_N)\right)\right) \right) \quad (10)$$

$$\text{s.t. } \mathbf{z} = E(\mathbf{x}); \mathbf{y}_1 = D_1(\mathbf{z}); \mathbf{y}_n = D_n(\mathbf{y}_{n-1}), n = 2,...,N;$$

$$C_z; \sum_{n=1}^{N} \|\mathbf{y}_n - \mathbf{x}\|_2^2 < \varepsilon, n = 1,...,N;$$

$$\mathbf{y} = \mathbf{y}_N$$

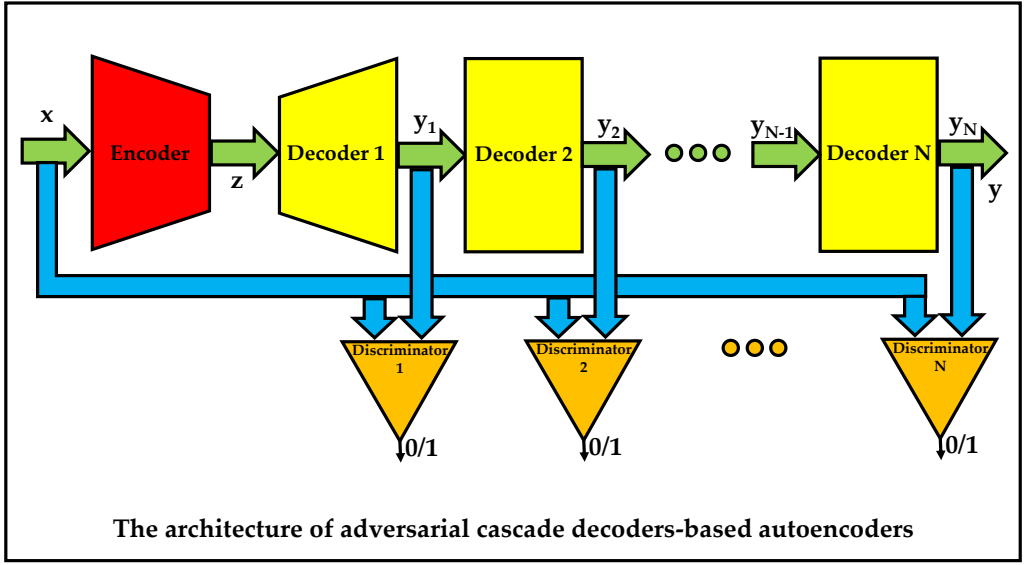

**The architecture of adversarial cascade decoders-based autoencoders**

**Figure 5.** The architecture of adversarial cascade decoders-based autoencoders.

3.3.3. Residual-adversarial cascade decoders-based autoencoders

The framework of residual-adversarial cascade decoders-based autoencoders (RACDAE) is shown in Fig. 6. The blue signal line denotes the training phase, and the green signal line denotes the both phases of training and testing. Each decoder is a residual-adversarial module.

The reconstruction data can be resolved by the following optimization problem:

$$(\theta; \mathbf{z}; \mathbf{r}_1, \cdots, \mathbf{r}_N; \mathbf{y}_1, \cdots, \mathbf{y}_N; \mathbf{y})$$
$$= \arg \min_{E;D} \max_{DC_1, \cdots, DC_N} \sum_{n=1}^{N} \left( \alpha_n M \left( \ln \left( DC_n \left( \mathbf{x} - \mathbf{y}_{n-1} \right) \right) \right) + \beta_n M \left( \ln \left( 1 - DC_n \left( \mathbf{r}_n \right) \right) \right) \right)$$
$$\text{s.t. } \mathbf{z} = E(\mathbf{x}); \mathbf{r}_1 = D_1(\mathbf{z}); \mathbf{r}_n = D_n(\mathbf{y}_{n-1}), n = 2, \ldots, N; \quad (11)$$
$$C_z; \sum_{n=1}^{N} \|\mathbf{x} - \mathbf{y}_{n-1} - \mathbf{r}_n\|_2^2 < \varepsilon, n = 1, \ldots, N;$$
$$\mathbf{y}_0 = \mathbf{0}; \mathbf{y}_n = \mathbf{y}_{n-1} + \mathbf{r}_n, n = 1, \ldots, N; \mathbf{y} = \mathbf{y}_N$$

For the purpose of gradually and serially training residual adversarial cascade decoders-based autoencoders, the optimization problem in Eq. 11 can be divided into the following sub optimization problems:

$$(\theta_1; \mathbf{z}; \mathbf{r}_1; \mathbf{y}_1) = \arg \min_{E;D_1} \max_{DC_1} \left( \alpha_1 M \left( \ln \left( DC_1 \left( \mathbf{x} - \mathbf{y}_0 \right) \right) \right) + \beta_1 M \left( \ln \left( 1 - DC_1 \left( \mathbf{r}_1 \right) \right) \right) \right)$$
$$(\theta_2; \mathbf{r}_2; \mathbf{y}_2) = \arg \min_{D_2} \max_{DC_2} \left( \alpha_2 M \left( \ln \left( DC_2 \left( \mathbf{x} - \mathbf{y}_1 \right) \right) \right) + \beta_2 M \left( \ln \left( 1 - DC_2 \left( \mathbf{r}_2 \right) \right) \right) \right)$$
$$\ldots\ldots$$
$$(\theta_N; \mathbf{r}_N; \mathbf{y}_N; \mathbf{y}) = \arg \min_{D_N} \max_{DC_N} \left( \alpha_N M \left( \ln \left( DC_N \left( \mathbf{x} - \mathbf{y}_{N-1} \right) \right) \right) + \beta_N M \left( \ln \left( 1 - DC_N \left( \mathbf{r}_N \right) \right) \right) \right) \quad (12)$$
$$\text{s.t. } \mathbf{z} = E(\mathbf{x}); \mathbf{r}_1 = D_1(\mathbf{z}); \mathbf{r}_n = D_n(\mathbf{y}_{n-1}), n = 2, \ldots, N;$$
$$C_z; \sum_{n=1}^{N} \|\mathbf{x} - \mathbf{y}_{n-1} - \mathbf{r}_n\|_2^2 < \varepsilon, n = 1, \ldots, N;$$
$$\mathbf{y}_0 = \mathbf{0}; \mathbf{y}_n = \mathbf{y}_{n-1} + \mathbf{r}_n, n = 1, \ldots, N; \mathbf{y} = \mathbf{y}_N$$

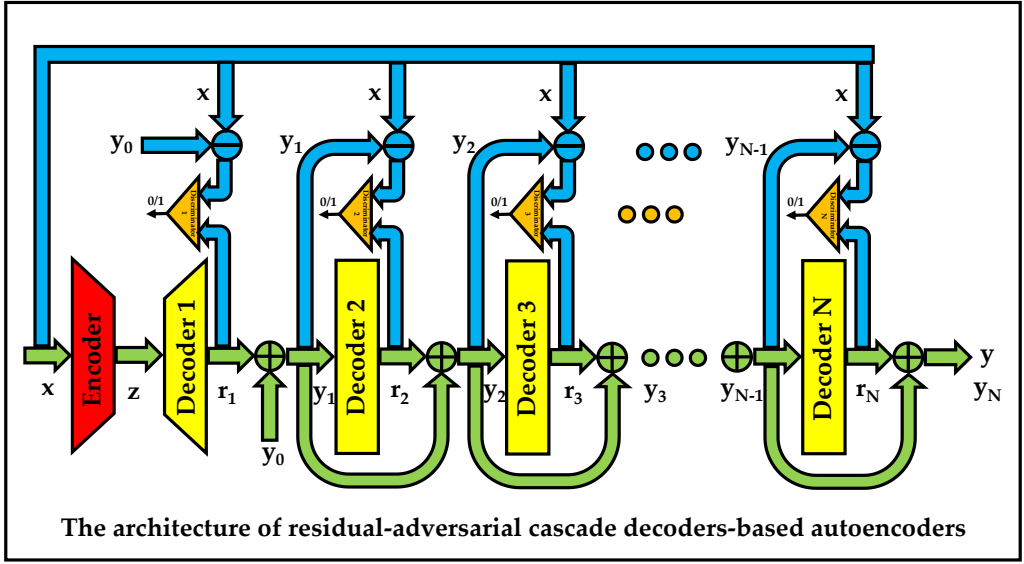

**Figure 6.** The architecture of residual-adversarial cascade decoders-based autoencoders.

*3.4. Adversarial autoencoders*

3.4.1. Reminiscence of classical adversarial autoencoders

The infrastructure of the classical adversarial autoencoders is exhibited in Fig. 7. The blue signal flow is for the training phase, and the green signal flow is for the both phases of training and testing. AAE are the combination of autoencoders and adversarial learning. Alireza Makhzani et al. propose the AAE, utilize the Encoder unit of autoen coders as generator and add an independent discriminator, employ adversarial learning in latent space and let the hidden variable satisfy a given distribution, and finally gain better performance of data reconstruction [13]. Compared with the classical autoencoders, the AAE infrastructure holds an extra discriminator, which makes the output of Encoder maximally approach a given distribution. The infrastructure can be expressed by the following equations:

$$\begin{aligned} \mathbf{z} &= E(\mathbf{x}) \\ \mathbf{y} &= D(\mathbf{z}) \\ DC(\mathbf{z}_h) &= 1, DC(\mathbf{z}) = 0 \end{aligned} \quad , \tag{13}$$

where:

$\mathbf{z}_h$ is the variable related to $\mathbf{z}$ which satisfies a given distribution;

DC is the Discriminator.

The reconstruction data can be resolved by the following optimization problem:

$$(\theta, \mathbf{z}, \mathbf{y}) = \arg \min_{E,D} \max_{DC} \left( \alpha M \left( \ln \left( DC(\mathbf{z}_h) \right) \right) + \beta M \left( \ln \left( 1 - DC(\mathbf{z}) \right) \right) \right) \\ \text{s.t. } \mathbf{z} = E(\mathbf{x}), \mathbf{y} = D(\mathbf{z}), \|\mathbf{y} - \mathbf{x}\|_2^2 < \varepsilon, C_y \tag{14}$$

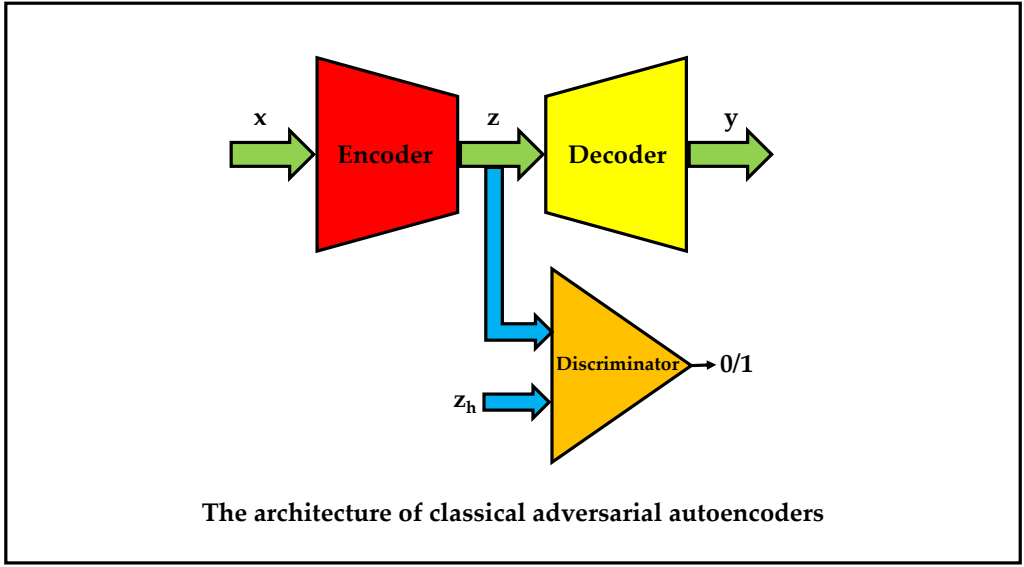

**The architecture of classical adversarial autoencoders**

**Figure 7.** The architecture of classical adversarial autoencoders.

3.4.2. Proposed cascade decoders-based adversarial autoencoders

The architecture of the proposed cascade decoders-based adversarial autoencoders (CDAAE) is illustrated in Fig. 8. The blue flow line represents the training phase, and the green flow line represents the both phases of training and testing. Compared with the cascade decoders-based autoencoders, the proposed architecture has an extra discriminator, which makes the output of Encoder maximally approximate to a known distribution. The architecture can be described by the following formulas:

$$\begin{aligned} \mathbf{z} &= E(\mathbf{x}) \\ \mathbf{y}_n &= \begin{cases} D_1(\mathbf{z}), n = 1 \\ D_n(\mathbf{y}_{n-1}), n = 2, ..., N \end{cases} , \\ DC(\mathbf{z}_h) &= 1, DC(\mathbf{z}) = 0 \end{aligned} \tag{15}$$

The reconstruction data can be resolved by the following optimization problem:

$$(\theta; \mathbf{z}; \mathbf{y}_1, \cdots, \mathbf{y}_N) = \arg \min_{E; D_1, \cdots, D_N} \max_{DC} \left( \alpha M \left( \ln \left( DC(\mathbf{z}_h) \right) \right) + \beta M \left( \ln \left( 1 - DC(\mathbf{z}) \right) \right) \right) \\ \text{s.t. } \mathbf{z} = E(\mathbf{x}); \mathbf{y}_1 = D_1(\mathbf{z}); \mathbf{y}_n = D_n(\mathbf{y}_{n-1}), n = 2, ..., N; \\ \sum_{n=1}^{N} \|\mathbf{y}_n - \mathbf{x}\|_2^2 < \varepsilon; C_{y_n}, n = 1, ..., N \tag{16}$$

For the purpose of gradually and serially training cascade Decoders-based adversarial autoencoders, the optimization problem in Eq. 16 can be partitioned into the following sub optimization problems:

$$\begin{aligned}
(\boldsymbol{\theta}_1; \mathbf{z}; \mathbf{y}_1) &= \arg\min_{E;D_1} \max_{DC} \left( \alpha M \left( \ln(DC(\mathbf{z}_h)) \right) + \beta M \ln(1 - DC(\mathbf{z})) \right) \\
(\boldsymbol{\theta}_2; \mathbf{y}_2) &= \arg\min_{D_2} \|\mathbf{y}_2 - \mathbf{x}\|_2^2 \\
&\ldots\ldots \\
(\boldsymbol{\theta}_N; \mathbf{y}_N) &= \arg\min_{D_N} \|\mathbf{y}_N - \mathbf{x}\|_2^2 \\
\text{s.t. } & \mathbf{z} = E(\mathbf{x}); \mathbf{y}_1 = D_1(\mathbf{z}); \mathbf{y}_n = D_n(\mathbf{y}_{n-1}), n = 2,\ldots,N; \\
& \|\mathbf{y}_1 - \mathbf{x}\|_2^2 < \varepsilon; C_{\mathbf{y}_n}, n = 1,\ldots,N
\end{aligned} \quad (17)$$

The architecture in Fig. 8 is general cascade decoders-based adversarial autoencoders (GCDAAE), and it can be easily be expanded to residual cascade decoders-based adversarial autoencoders (RCDAAE), adversarial cascade decoders-based adversarial autoencoders (ACDAAE) and residual-adversarial cascade decoders-based adversarial autoencoders (RACDAAE).

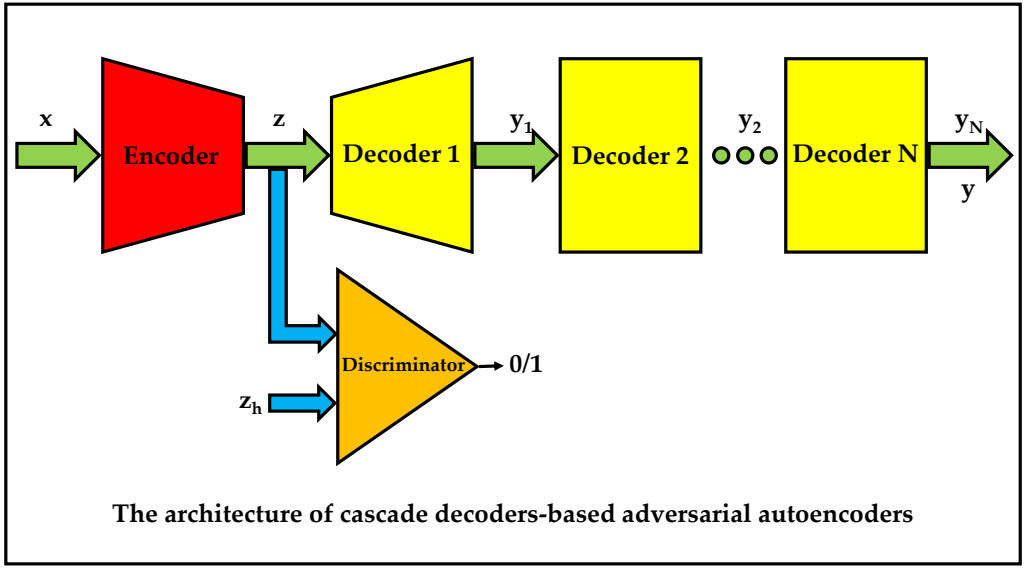

**Figure 8.** The architecture of cascade decoders-based adversarial autoencoders.

*3.5. Variational autoencoders*

3.5.1. Remembrance of classical variational autoencoders

The framework of classical variational autoencoders is shown in Fig. 9 [4]. The blue signal line denotes the training phase, and the green signal line denotes the both phases of training and testing. It can be resolved by the following optimization problem:

$$\begin{aligned}
(\boldsymbol{\theta}, \mathbf{z}, \mathbf{y}) &= \arg\min_{\boldsymbol{\theta},\mathbf{z},\mathbf{y}} \left( \alpha \mathrm{KL}\left( q(\mathbf{z}_h) \| p(\mathbf{z}) \right) + \beta \|\mathbf{y} - \mathbf{x}\|_2^2 \right) \\
&= \arg\min_{\boldsymbol{\theta},\mathbf{z},\mathbf{y}} \left( \alpha \sum_{\mathbf{z}_h} q(\mathbf{z}_h) \log \frac{q(\mathbf{z}_h)}{p(\mathbf{z})} + \beta \|\mathbf{y} - \mathbf{x}\|_2^2 \right) \\
\text{s.t. } & \mathbf{z} = E(\mathbf{x}), \mathbf{y} = D(\mathbf{z}), C_y
\end{aligned} \quad (18)$$

where:

KL(·) is the Kullback-Leibler divergence;

q($\mathbf{z}_h$) is the given distribution of $\mathbf{z}_h$;

p($\mathbf{z}$) is the distribution of $\mathbf{z}$.

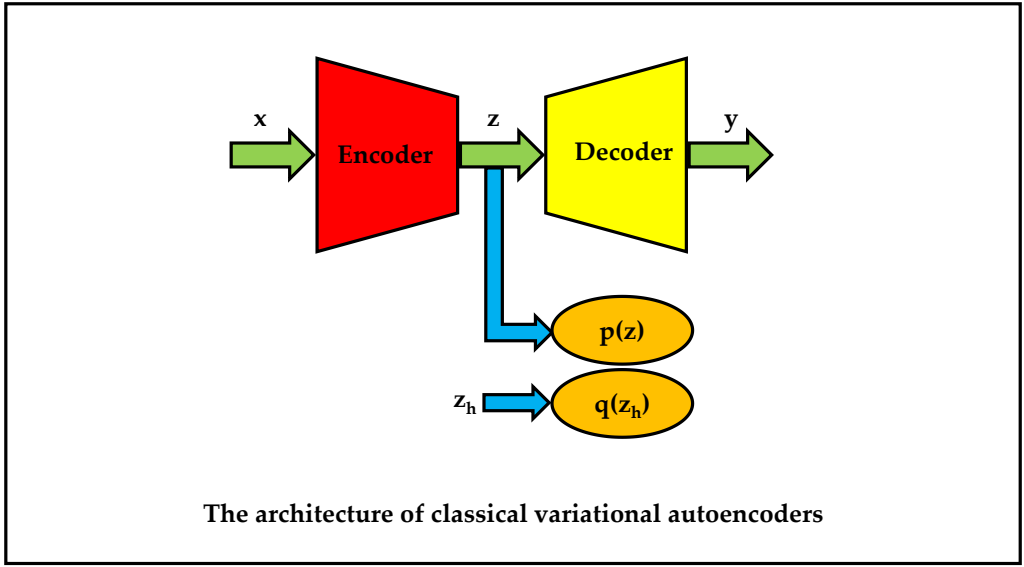

**Figure 9.** The architecture of classical variational autoencoders.

3.5.2. Proposed cascade decoders-based variational autoencoders

The proposed infrastructure of cascade decoders-based variational autoencoders is shown in Fig. 10. The blue signal flow is for the training phase, and the green signal flow is for the both phases of training and testing. It can be resolved by the following optimization problem:

$$(\theta; \mathbf{z}; \mathbf{y}_1, \cdots, \mathbf{y}_N) = \underset{\theta; \mathbf{z}; \mathbf{y}_1, \cdots, \mathbf{y}_N}{\arg\min}\left(\alpha \mathrm{KL}(q(\mathbf{z}_h) || p(\mathbf{z})) + \beta \sum_{n=1}^{N} \|\mathbf{y}_n - \mathbf{x}\|_2^2 \right), \quad (19)$$

$$\text{s.t. } \mathbf{z} = E(\mathbf{x}); \mathbf{y}_1 = D_1(\mathbf{z}); \mathbf{y}_n = D_n(\mathbf{y}_{n-1}), n = 2, ..., N; C_{\mathbf{y}_n}, n = 1, ..., N$$

For the sake of gradually and serially training cascade decoders-based autoencoders, the optimization problem in Eq. 19 can be divided into the following sub optimization problems:

$$(\theta_1; \mathbf{z}; \mathbf{y}_1) = \underset{\theta_1; \mathbf{z}; \mathbf{y}_1}{\arg\min}\left(\alpha \mathrm{KL}(q(\mathbf{z}_h) || p(\mathbf{z})) + \beta \|\mathbf{y}_1 - \mathbf{x}\|_2^2 \right)$$

$$(\theta_2; \mathbf{y}_2) = \underset{\theta_2; \mathbf{y}_2}{\arg\min} \|\mathbf{y}_2 - \mathbf{x}\|_2^2$$

$$\ldots\ldots \quad , \quad (20)$$

$$(\theta_N; \mathbf{y}_N) = \underset{\theta_N; \mathbf{y}_N}{\arg\min} \|\mathbf{y}_N - \mathbf{x}\|_2^2$$

$$\text{s.t. } \mathbf{z} = E(\mathbf{x}); \mathbf{y}_1 = D_1(\mathbf{z}); \mathbf{y}_n = D_n(\mathbf{y}_{n-1}), n = 2, ..., N; C_{\mathbf{y}_n}, n = 1, ..., N$$

The infrastructure in Fig. 10 is general cascade decoders-based variational autoencoders (GCDVAE), and it can be easily be extended to residual cascade decoders-based variational autoencoders (RCDVAE), adversarial cascade decoders-based variational autoencoders (ACDVAE) and residual-adversarial cascade decoders-based variational autoencoders (RACDVAE).

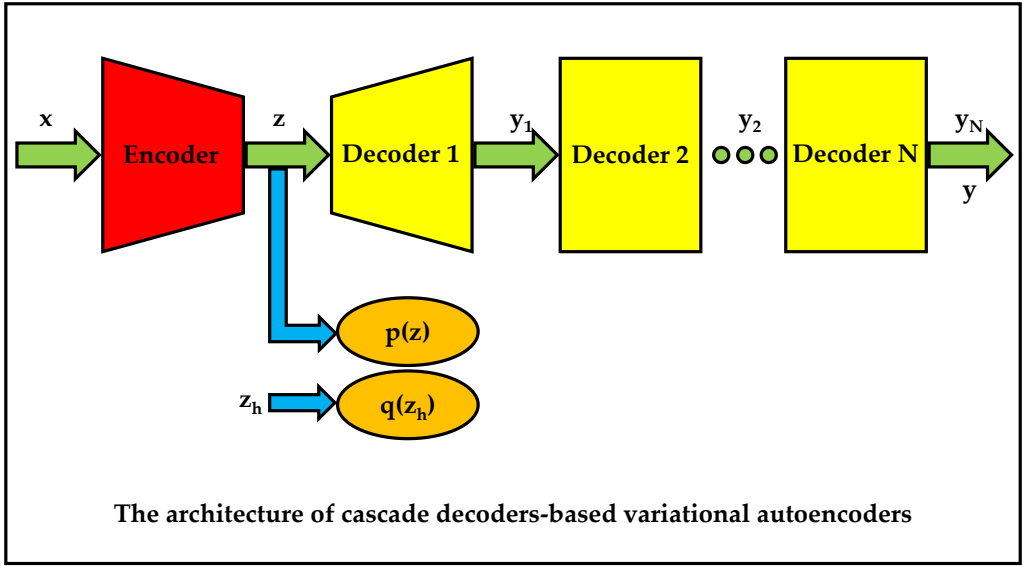

**Figure 10.** The architecture of cascade decoders-based variational autoencoders.

*3.6. Wasserstein autoencoders*

3.6.1. Recollection of classical Wasserstein autoencoders

The classical Wasserstein autoencoders can be resolved by the following optimization problem [20]:

$$(\theta, \mathbf{z}, \mathbf{y}) = \underset{\theta, \mathbf{z}, \mathbf{y}}{\arg\min} \left( \alpha W_z \left( p(\mathbf{z}), q(\mathbf{z}_h) \right) + \beta W_y \left( \mathbf{y}, \mathbf{x} \right) \right),$$
$$\text{s.t. } \mathbf{z} = E(\mathbf{x}), \mathbf{y} = D(\mathbf{z}), C_y \qquad (21)$$

where:

$W_z$ is the regularizer between distribution p(**z**) and $q(\mathbf{z}_h)$;

$W_y$ is the reconstruction cost.

3.6.2. Proposed cascade decoders-based Wasserstein autoencoders

The proposed cascade decoders-based Wasserstein autoencoders can be resolved by the following optimization problem:

$$(\theta; \mathbf{z}; \mathbf{y}_1, \cdots, \mathbf{y}_N) = \underset{\theta; \mathbf{z}; \mathbf{y}_1, \cdots, \mathbf{y}_N}{\arg\min} \left( \alpha W_z \left( p(\mathbf{z}), q(\mathbf{z}_h) \right) + \beta \sum_{n=1}^{N} W_y \left( \mathbf{y}_n, \mathbf{x} \right) \right),$$
$$\text{s.t. } \mathbf{z} = E(\mathbf{x}); \mathbf{y}_1 = D_1(\mathbf{z}); \mathbf{y}_n = D_n(\mathbf{y}_{n-1}), n = 2, \ldots, N; C_{y_n}, n = 1, \ldots, N \qquad (22)$$

For the purpose of gradually and serially training cascade decoders-based autoencoders, the optimization problem in Eq. 22 can be divided into the following sub optimization problems:

$$(\theta_1; \mathbf{z}; \mathbf{y}_1) = \underset{\theta_1; \mathbf{z}; \mathbf{y}_1}{\arg\min} \left( \alpha W_z \left( p(\mathbf{z}), q(\mathbf{z}_h) \right) + \beta W_y \left( \mathbf{y}_1, \mathbf{x} \right) \right)$$
$$(\theta_2; \mathbf{y}_2) = \underset{\theta_2; \mathbf{y}_2}{\arg\min} W_y \left( \mathbf{y}_2, \mathbf{x} \right)$$
$$\ldots\ldots$$
$$(\theta_N; \mathbf{y}_N) = \underset{\theta_N; \mathbf{y}_N}{\arg\min} W_y \left( \mathbf{y}_N, \mathbf{x} \right)$$
$$\text{s.t. } \mathbf{z} = E(\mathbf{x}); \mathbf{y}_1 = D_1(\mathbf{z}); \mathbf{y}_n = D_n(\mathbf{y}_{n-1}), n = 2, \ldots, N; C_{y_n}, n = 1, \ldots, N \qquad (23)$$

The aforementioned architecture is general cascade decoders-based Wasserstein autoencoders (GCDWAE), and it can be easily be expanded to residual cascade decoders-based Wasserstein autoencoders (RCDWAE), adversarial cascade decoders-based Wasserstein autoencoders (ACDWAE) and residual-adversarial cascade decoders-based Wasserstein autoencoders (RACDWAE).

*3.7. Pseudocodes of cascade decoders-based autoencoders*

The pseudocodes of the proposed cascade decoders-based autoencoders are shown in Fig. 11.

---

**Algorithm**: Cascade Decoders-Based Autoencoders
**Input**: **x**: the training data
      I: the total number of iteration
      N: the total number of sub minimization problem
**Initialization**: i=1
**Training**:
   While i<=I
      i++
      n=1
      While n<=N
         n++
         resolve the nth sub problem in Eq. (5), (7), (10), (12), (17), (20) or (23)
**Output**:
   θ: the parameters of deep neural networks
   **z**: the representations of hidden space
   $\mathbf{y}_1,\ldots,\mathbf{y}_N$: the output of cascade decoders
   **y**: the output of the last decoder

---

**Figure 11.** The pseudocodes of cascade decoders-based autoencoders.

## 4. Experiments

*4.1. Experimental datasets*

The purpose of the simulation experiments is to compare the data reconstruction performance of the proposed cascade decoders-based autoencoders and the classical autoencoders.

Four datasets are utilized to evaluate the algorithm performance [29-32]. The mixed national institute of standards and technology (MNIST) dataset has 10 classes of handwritten digit images [29], the extending MNIST (EMNIST) dataset holds 6 subcategories of handwritten digit and letter images [30], the fashion-MNIST (FMNIST) dataset possesses 10 classes of fashion product images [31], and the medical MNIST (MMNIST) dataset owns 10 subcategories of medical images [32]. The image size is 28x28. All color images are converted into gray images. In order to reduce the computation load, small resolution images and gray images are chosen. Certainly, if the computation capability is ensured, the proposed methods can be easily and directly utilized on big resolution images, the components of color images or their sub patches. A large image can be divided into small patches. In traditional image compression methods, the size of image patch for compression is 8x8. Therefore, the proposed methods can be used for each image block. In brief, large image size will not degrade the performance of the proposed methods from viewpoint of small image patches. For the convenience of training and testing deep neural networks, each pixel value is normalized from range [0,255] to range [-1,+1] in the phase of preprocessing and is rescaled back to range [0,255] in the phase of post processing. The numbers of classes and samples in the four datasets are enumerated in Tab. 2. The sample images of the four datasets is illustrated in Fig. 12. From top to bottom, there are images of MNIST digits, EMINST digits, EMNIST letters, FMNIST goods, MMNIST breast, chest, derma, optical coherence tomography (OCT), axial organ, coronal organ, sagittal organ, pathology, pneumonia, and retina.

*4.2. Experimental conditions*

The components of autoencoders make up of fully-connected (FC) layers, leaky rectified linear unit (LRELU) layers, hyperbolic tangent (Tanh) layers, Sigmoid layers, and etc. In order to reduce the calculation complexity, the convolutional (CONV) layer is not utilized.

The composition of Encoder is shown in Fig. 13, which consists of input, FC, LRELU and hidden layers. The constitution of Decoder is illustrated in Fig. 14, which consists of input, FC, LRELU, Tanh and output layers. The input layer can be the hidden layer for the first Decoder and be the output layer of preceding Decoder for the latter Decoders. The dashed line shows the two situations.

The organization of discriminator is demonstrated in Fig. 15, which comprises input, FC, LRELU, Sigmoid and output layers. The input layer can be the hidden layer and be the output of each Decoder. The dashed line indicates the two cases.

The deep learning parameters, such as image size, latent dimension, decoder number, batch size, learning rate and iteration epoch, are summarized in the following Tab. 3.

*4.3. Experimental results*

The experimental results of the proposed and classical algorithms on MNIST dataset, EMNIST dataset, FMNIST dataset and MMNIST dataset are respectively shown in Tab. 4-7. SSIM is the average structure similarity between reconstruction images and original images. ΔSIMM is the average SSIM difference between the proposed approaches and the conventional AE approach. The experimental results are also displayed in Fig. 16, where the horizontal coordinate is datasets and the vertical coordinate is ΔSIMM.

**Table 2.** Class and sample numbers of experimental datasets.

| Dataset | | Class Number | Sample Number | |
| --- | --- | --- | --- | --- |
| | | | Training | Testing |
| **MNIST** | | 10 | 60000 | 10000 |
| **EMINST** | digits | 10 | 240000 | 40000 |
| | letters | 26 | 124800 | 20800 |
| | balanced | 47 | 112800 | 18800 |
| | bymerge | 47 | 697932 | 116323 |
| | byclass | 62 | 697932 | 116323 |
| **FMNIST** | | 10 | 60000 | 10000 |
| **MMNIST** | breast | 2 | 546 | 156 |
| | chest | 2 | 78468 | 22433 |
| | derma | 7 | 7007 | 2005 |
| | OCT | 4 | 97477 | 1000 |
| | axial organ | 11 | 34581 | 17778 |
| | coronal organ | 11 | 13000 | 8268 |
| | sagittal organ | 11 | 13940 | 8829 |
| | pathology | 9 | 89996 | 7180 |
| | pneumonia | 2 | 4708 | 624 |
| | retina | 5 | 1080 | 400 |

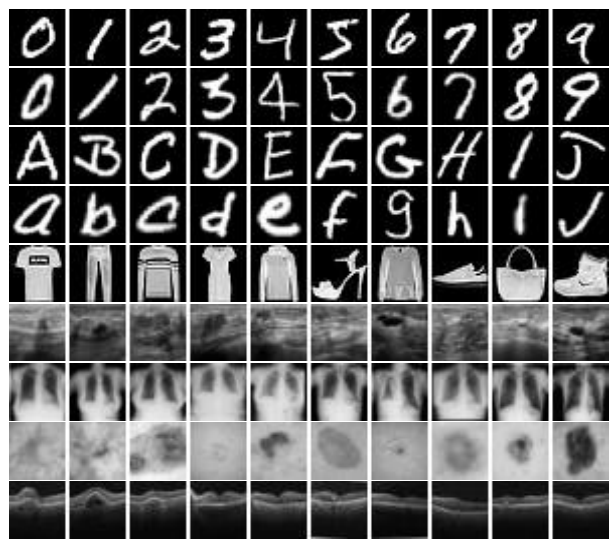

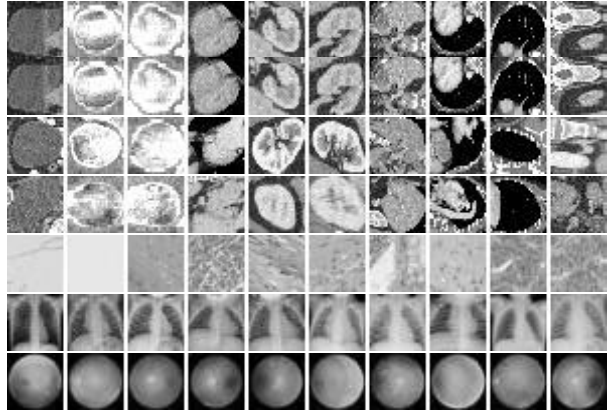

**Figure 12.** Sample images of experimental datasets.

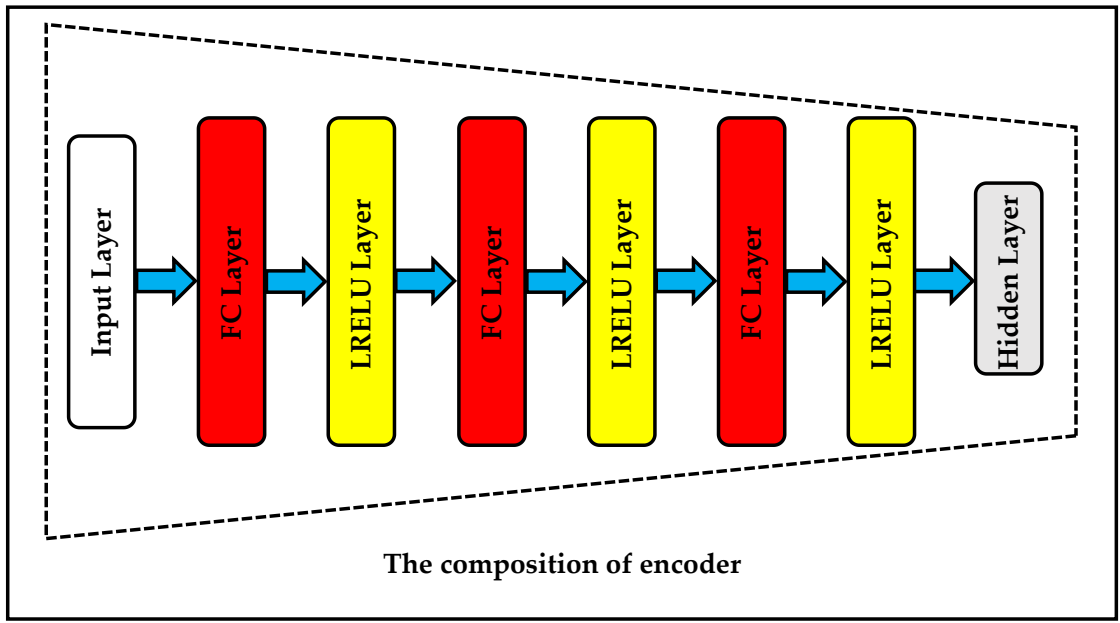

**Figure 13.** The composition of Encoder.

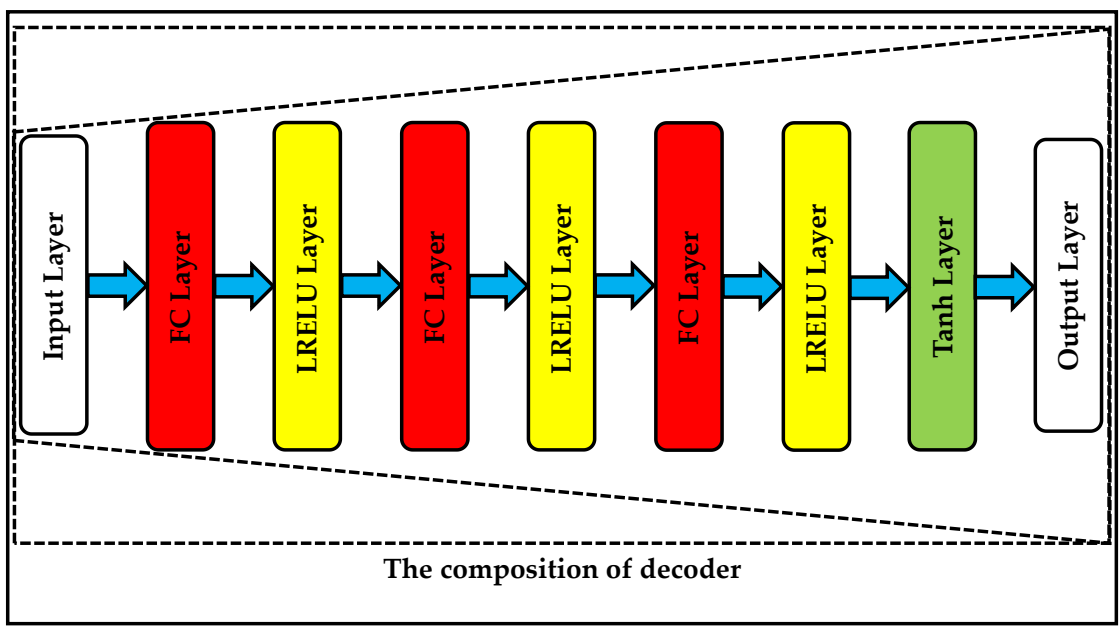

**Figure 14.** The composition of Decoder.

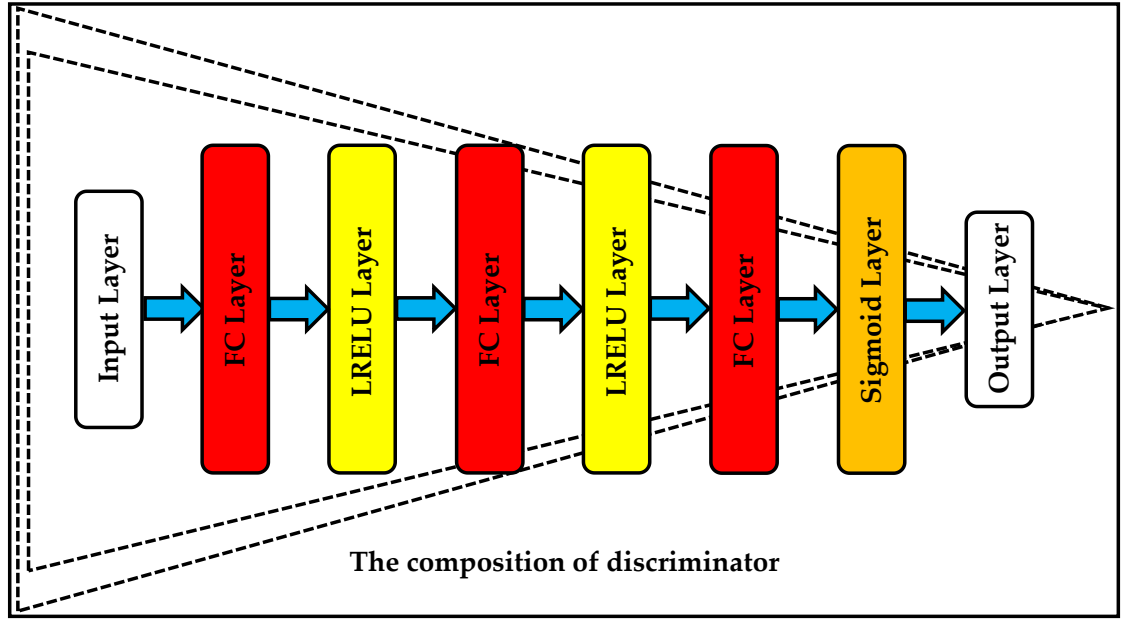

**Figure 15.** The composition of discriminator.

**Table 3.** The deep learning parameters.

| Parameter Name | Parameter Value |
|---|---|
| Image size | 28x28x1 |
| Latent Dimension | 30 |
| Decoder Number | 3 |
| Batch size | 100 |
| Learning Rate | 0.0002 |
| Iteration Epoch | 100 |

It can be found in Tab. 4-7 and Fig.16 that the proposed methods, except for ACDAE and ACDAAE, are superior to the classical AE and AAE methods in the performance of image reconstruction. Therefore, it proves the correctness and effectiveness of the proposed cascade decoders-based autoencoders for image reconstruction.

It can also be discovered in Tab. 4-7 and Fig.16 that the proposed RCDAE and RACDAAE algorithms almost hold the best recovery performance on the four datasets. Hence, residual learning is very suitable for image recovery. This is owing to the fact that residual has smaller average and variance than the original image and is beneficial for the deep neural network to learn the relationship between input and output.

It can further be sought in Tab. 4-7 and Fig.16 that the proposed ACDAE and ACDAAE algorithms hold some minus ΔSIMM on the four datasets. Thus, under the condition of this paper, pure adversarial learning is unsuitable for image reestablishment. However, the combination of residual learning and adversarial learning, such as aforementioned RACDAAE, can obtain high reestablishment performance.

It can additionally be found in Tab. 4-7 and Fig.16 that the AAE algorithm possesses some minus ΔSIMM on the four datasets. Therefore, under the circumstance of this article, pure AAE cannot outperform AE in image reconstitution. Nevertheless, the combination of residual learning and adversarial learning, such as aforementioned RACDAAE, can gain high reconstitution performance.
It can finally be found in Tab. 4-7 and Fig.16 that SSIM difference for MMNIST-axial and MMNIST-sagittal is very higher than other data sets. The reason may be that the training and testing samples are more similar than other datasets.

In order to clearly compare the reconstruction performance between the proposed algorithms and the classical algorithms, the reconstruction images are illustrated in Fig. 17-26. Because the proposed RCDAE algorithm owns the best performance, it is taken as an example.

The recovery images on MNIST dataset are shown in Fig. 17. For each sub image in Fig. 17, the top row is the original images, the middle row is the recovery images of AE, and the bottom row is the recovery images of RCDAE. It is uneasy to find the SSIM difference between AE and RCDAE in Fig. 17. Therefore, the marked reestablishment images on MNIST dataset are illustrated in Fig. 18. The left column is the original images, the middle column is the reestablishment images of AE, and the right column is the reestablishment images of RCDAE. It is easy to check the SSIM difference between AE and RCDAE in the red marked squares in Fig. 18.

Similarly, the reconstitution images on EMNIST dataset (big letters) are demonstrated in Fig. 19; The marked reconstitution images on EMNIST dataset (big letters) are demonstrated in Fig. 20. The rebuilding images on EMNIST dataset (small letters) are displayed in Fig. 21; The marked rebuilding images on EMNIST dataset (small letters) are displayed in Fig. 22. The reconstruction images on FMNIST dataset are shown in Fig. 23; The marked reconstruction images on FMNIST dataset are shown in Fig. 24. The recovery images on MMNIST dataset are displayed in Fig. 25; The marked recovery images on MMNIST dataset are displayed in Fig. 26.

It is revealed in Fig. 17-26, the proposed algorithms achieve significant improvement of reestablishment performance on MNST and EMNIST datasets. It is also manifested in Fig. 17-26, the proposed methods merely obtain unobvious promotion of reestablishment performance on FMNIST and MMIST datasets. For instance, in the first row of Fig. 26, the difference between the proposed and classical methods can only be found after enlarging the images; in the eighth row of Fig.26, the conspicuous difference still cannot be found even after enlarging the images. Nevertheless, both of them are the true experimental results, which should be accepted and explained. This is due to the following four reasons. The first reason is that the quality of the original images is low on FMNIST and MMNIST datasets. The second reason is that only illumination component of original color image on part of MMNIST datasets is reserved. The reconstruction performance will be improved if the original color image is utilized. The third reason is that the dimension of laten space is 30. It a very low choice compared with 784 (28x28), the dimension of the original image. The fourth reason is that convolutional layer is not utilized in the architecture of autoencoders. For the purpose of decreasing the computation load, the convolutional layer is not adopted in the proposed approaches. Convolutional layer can effectively extract image features and reconstruct original image, is expected to further improve the reconstruction performance of the proposed approaches, and it will be investigated in our future work.

## 5. Conclusions

This paper proposes cascade decoders-based autoencoders for image reconstruction. They comprise the architecture of multi-level decoders and related optimization problems and training algorithms. This article concentrates on the classical AE and AAE, and their serial decoders-based versions. Residual learning and adversarial learning are contained in the proposed approaches. The effectiveness of cascade decoders for image reconstruction is demonstrated in mathematics. It is evaluated by the experimental results on four open datasets that the proposed cascade decoders-based autoencoders are superior to the classical autoencoders in the performance of image reconstruction. Especially, residual learning is very fit for image reconstruction.

**Table 4.** Experimental results on MNIST dataset.

| Algorithms | SSIM | ΔSSIM |
|---|---|---|
| AE | 0.97387 | 0.00000 |
| GCDAE | 0.97415 | 0.00028 |
| **RCDAE** | **0.97592** | **0.00205** |
| ACDAE | 0.97354 | -0.00033 |
| RACDAE | 0.97574 | 0.00187 |
| AAE | 0.97304 | -0.00083 |
| GCDAAE | 0.97397 | 0.00010 |
| RCDAAE | 0.97576 | 0.00189 |
| ACDAAE | 0.97381 | -0.00006 |
| RACDAAE | 0.97589 | 0.00202 |

**Table 5.** Experimental results on EMNIST dataset.

| Algorithms | SSIM ΔSSIM EMNIST | | | | |
|---|---|---|---|---|---|
| | digits | letters | balanced | bymerge | byclass |
| AE | 0.98322 | 0.97466 | 0.97277 | 0.98288 | 0.98309 |
| | 0.00000 | 0.00000 | 0.00000 | 0.00000 | 0.00000 |
| GCDAE | 0.98380 | 0.97466 | 0.97315 | 0.98423 | 0.98432 |
| | 0.00058 | 0.00000 | 0.00038 | 0.00135 | 0.00123 |
| RCDAE | 0.98528 | 0.97672 | 0.97528 | **0.98589** | **0.98597** |
| | 0.00206 | 0.00206 | 0.00251 | **0.00301** | **0.00288** |
| ACDAE | 0.98285 | 0.97391 | 0.97223 | 0.98300 | 0.98314 |
| | -0.00037 | -0.00075 | -0.00054 | 0.00012 | 0.00005 |
| RACDAE | 0.98517 | 0.97654 | 0.97570 | 0.98433 | 0.98517 |
| | 0.00195 | 0.00188 | 0.00293 | 0.00145 | 0.00208 |
| AAE | 0.98304 | 0.97434 | 0.97291 | 0.98291 | 0.98284 |
| | -0.00018 | -0.00032 | 0.00014 | 0.00003 | -0.00025 |
| GCDAAE | 0.98375 | 0.97451 | 0.97305 | 0.98413 | 0.98476 |
| | 0.00053 | -0.00015 | 0.00028 | 0.00125 | 0.00167 |
| RCDAAE | 0.98534 | 0.97659 | 0.97546 | 0.98586 | 0.98592 |
| | 0.00212 | 0.00193 | 0.00269 | 0.00298 | 0.00283 |
| ACDAAE | 0.98305 | 0.97410 | 0.97275 | 0.98329 | 0.98340 |
| | -0.00017 | -0.00056 | -0.00020 | 0.00041 | 0.00031 |
| RACDAAE | **0.98543** | **0.97708** | **0.97593** | 0.98529 | 0.98531 |
| | **0.00221** | **0.00242** | **0.00316** | 0.00241 | 0.00222 |

**Table 6.** Experimental results on FMNIST dataset.

| Algorithms | SSIM | ΔSSIM |
|---|---|---|
| AE | 0.96335 | 0.00000 |
| GCDAE | 0.96463 | 0.00128 |
| RCDAE | 0.96620 | 0.00285 |
| ACDAE | 0.96329 | -0.00006 |
| RACDAE | 0.96625 | 0.00290 |
| AAE | 0.96366 | 0.00031 |
| GCDAAE | 0.96458 | 0.00123 |
| RCDAAE | 0.96630 | 0.00295 |
| ACDAAE | 0.96353 | 0.00018 |
| **RACDAAE** | **0.96637** | **0.00302** |

**Table 7.** Experimental results on MMNIST dataset.

| Algorithms | SSIM ΔSSIM MedMNIST | | | | | | | | | |
|---|---|---|---|---|---|---|---|---|---|---|
| | breast | chest | derma | OCT | axial | coronal | sagittal | pathology | pneumonia | retina |
| AE | 0.88868 | 0.98363 | 0.97103 | 0.98851 | 0.81806 | 0.82470 | 0.79883 | 0.89644 | 0.94760 | 0.97366 |
| | 0.00000 | 0.00000 | 0.00000 | 0.00000 | 0.00000 | 0.00000 | 0.00000 | 0.00000 | 0.00000 | 0.00000 |
| GCDAE | 0.89228 | 0.98836 | 0.97103 | 0.98856 | 0.92817 | 0.83320 | 0.84211 | 0.89720 | 0.95782 | 0.97751 |
| | 0.00360 | 0.00473 | 0.00000 | 0.00005 | 0.11011 | 0.00850 | 0.04328 | 0.00076 | 0.01022 | 0.00385 |
| RCDAE | **0.90574** | **0.98857** | **0.97396** | **0.98927** | 0.90092 | 0.83467 | 0.84180 | 0.89724 | **0.96115** | 0.98233 |
| | **0.01706** | **0.00494** | **0.00295** | **0.00076** | 0.08286 | 0.00997 | 0.04297 | 0.00080 | **0.01355** | 0.00867 |
| ACDAE | 0.87883 | 0.98682 | 0.96834 | 0.98780 | 0.93012 | 0.83450 | **0.84302** | 0.89548 | 0.95469 | 0.97892 |
| | -0.00985 | 0.00319 | -0.00269 | -0.00071 | 0.11206 | 0.00980 | **0.04419** | -0.00096 | 0.00709 | 0.00526 |
| RACDAE | 0.89908 | 0.98543 | 0.97028 | 0.98832 | 0.89931 | **0.83494** | 0.84155 | 0.89594 | 0.95784 | 0.98067 |
| | 0.01040 | 0.00180 | -0.00075 | -0.00019 | 0.08125 | **0.01024** | 0.04272 | -0.00050 | 0.01008 | 0.00701 |
| AAE | 0.88527 | 0.97240 | 0.97052 | 0.98858 | 0.81498 | 0.83347 | 0.84202 | 0.89603 | 0.95356 | 0.97646 |
| | -0.00341 | -0.01230 | -0.00051 | 0.00007 | -0.00308 | 0.00877 | 0.04319 | -0.00041 | 0.00596 | 0.00280 |
| GCDAAE | 0.89139 | 0.98847 | 0.97054 | 0.98867 | **0.93736** | 0.83347 | 0.84261 | 0.89714 | 0.95811 | 0.97839 |
| | 0.00271 | 0.00484 | -0.00053 | 0.00016 | **0.11930** | 0.01000 | 0.04378 | 0.00070 | 0.01051 | 0.00473 |
| RCDAAE | 0.89772 | 0.98852 | 0.97262 | 0.98920 | 0.89873 | 0.83321 | 0.84179 | **0.89747** | 0.96076 | **0.98261** |
| | 0.00094 | 0.00489 | 0.00159 | 0.00069 | 0.08607 | 0.00851 | 0.04296 | **0.00103** | 0.01316 | **0.00895** |
| ACDAAE | 0.88895 | 0.98673 | 0.96987 | 0.98769 | 0.93286 | 0.83421 | 0.84129 | 0.89695 | 0.95389 | 0.97942 |
| | 0.00027 | 0.00310 | -0.00116 | -0.00082 | 0.11480 | 0.00951 | 0.04246 | 0.00051 | 0.00629 | 0.00576 |
| RACDAAE | 0.90113 | 0.98831 | 0.97062 | 0.98865 | 0.90580 | 0.83380 | 0.83975 | 0.89728 | 0.95720 | 0.98117 |
| | 0.01245 | 0.00468 | -0.00041 | 0.00014 | 0.08702 | -0.00090 | 0.04092 | 0.00084 | 0.00510 | 0.00751 |

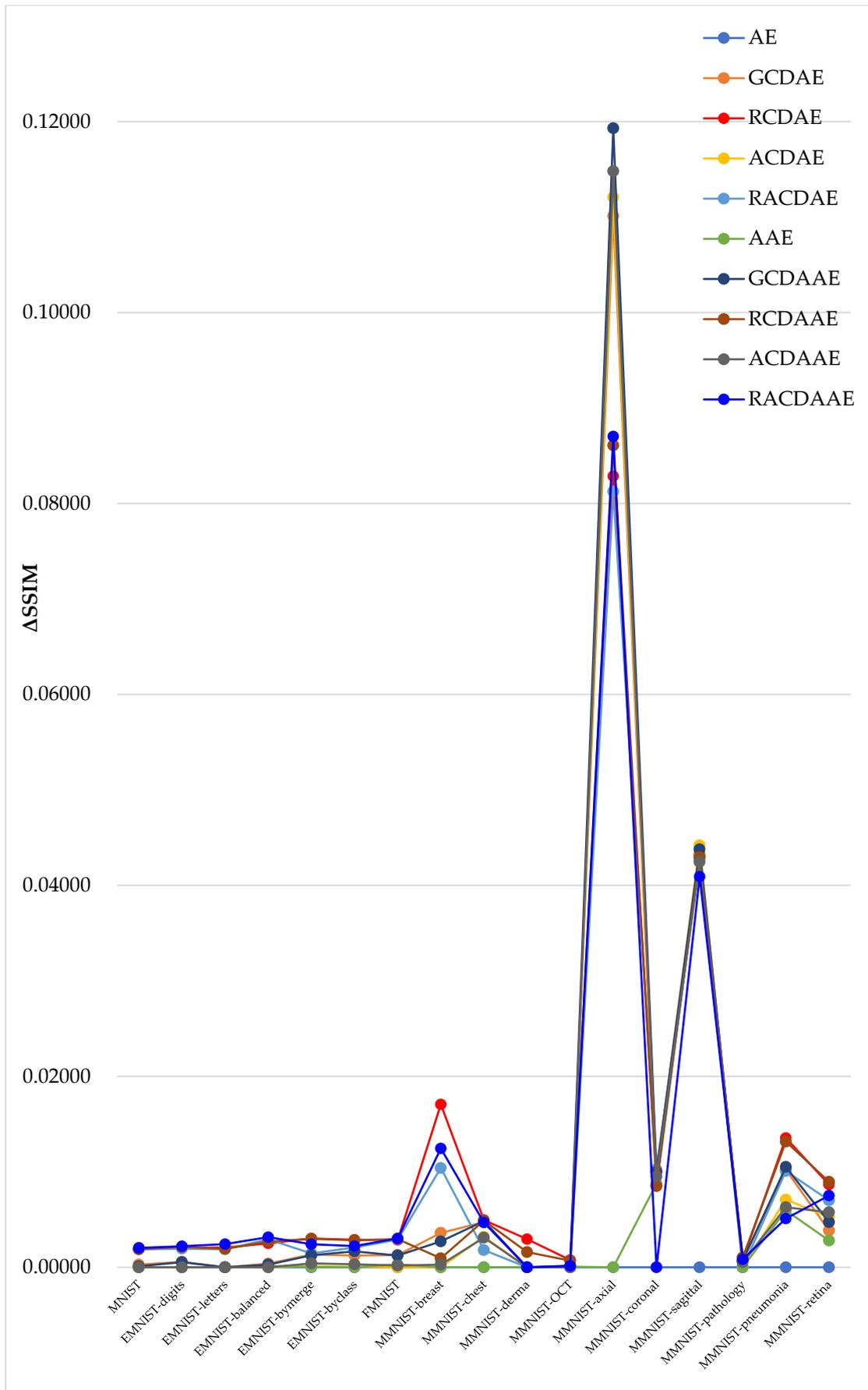

**Figure 16.** The experimental results of different algorithms on four datasets.

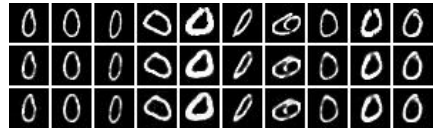

(1)

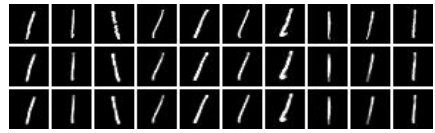

(2)

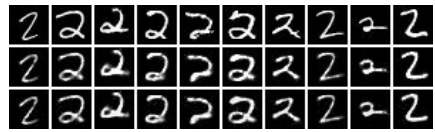

(3)

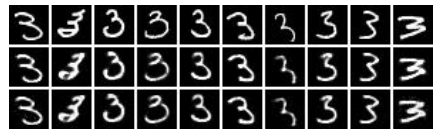

(4)

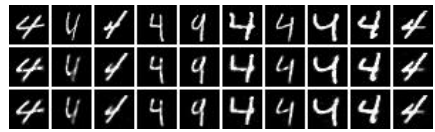

(5)

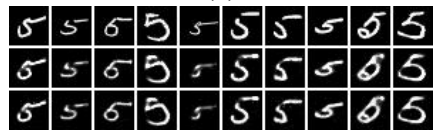

(6)

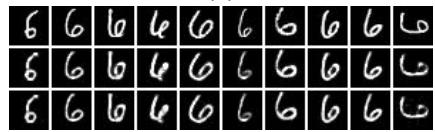

(7)

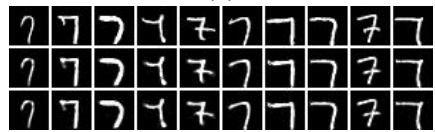

(8)

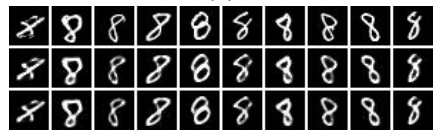

(9)

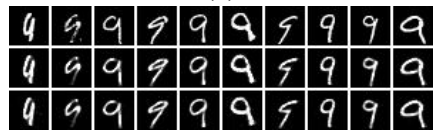

(10)

**Figure 17.** Reconstruction images on MNIST dataset.

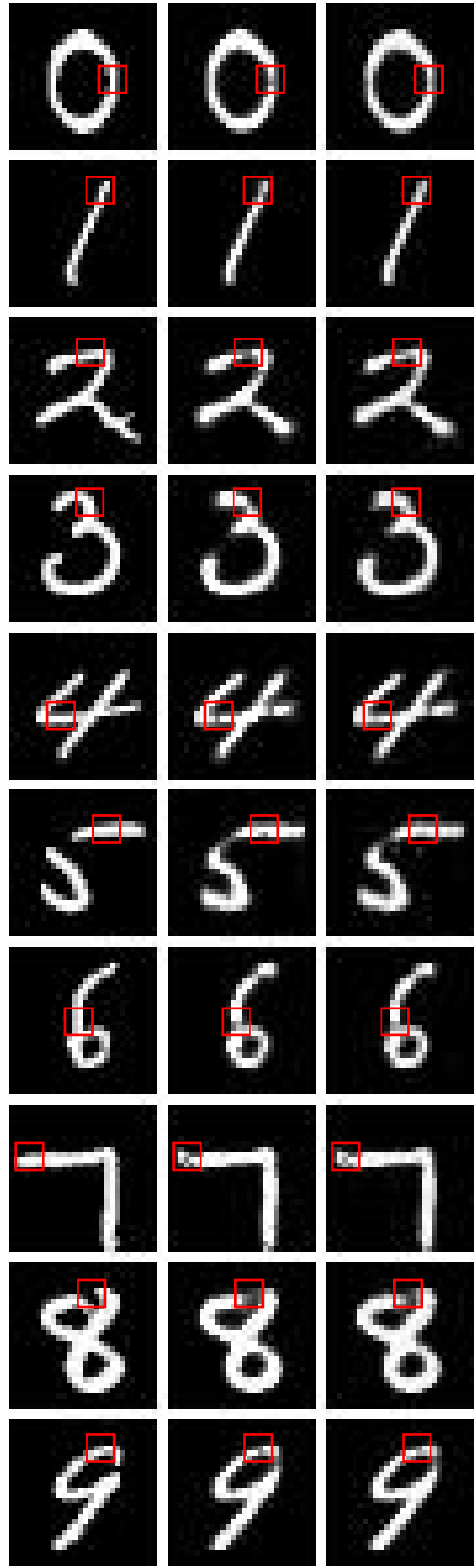

**Figure 18.** Marked reconstruction images on MNIST dataset.

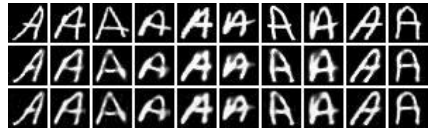
(1)
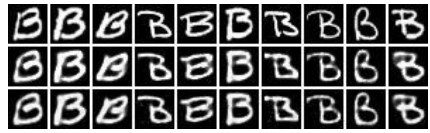
(2)
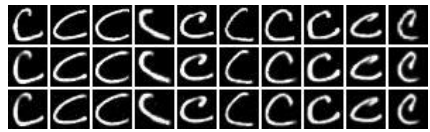
(3)
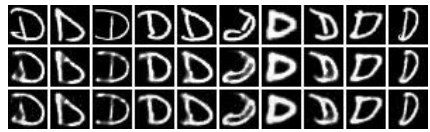
(4)
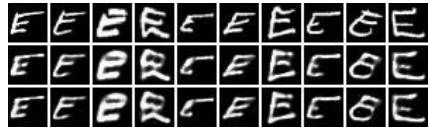
(5)
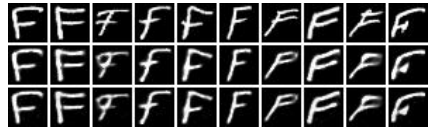
(6)
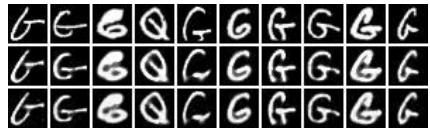
(7)
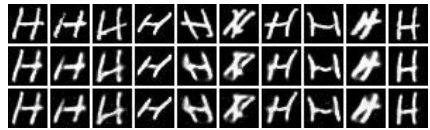
(8)
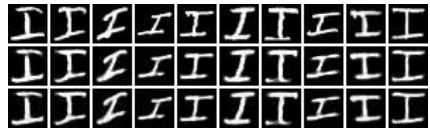
(9)
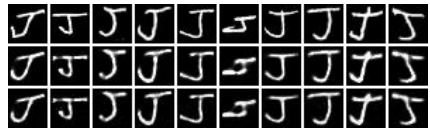
(10)

**Figure 19.** Reconstruction image on EMNIST dataset (big letters).

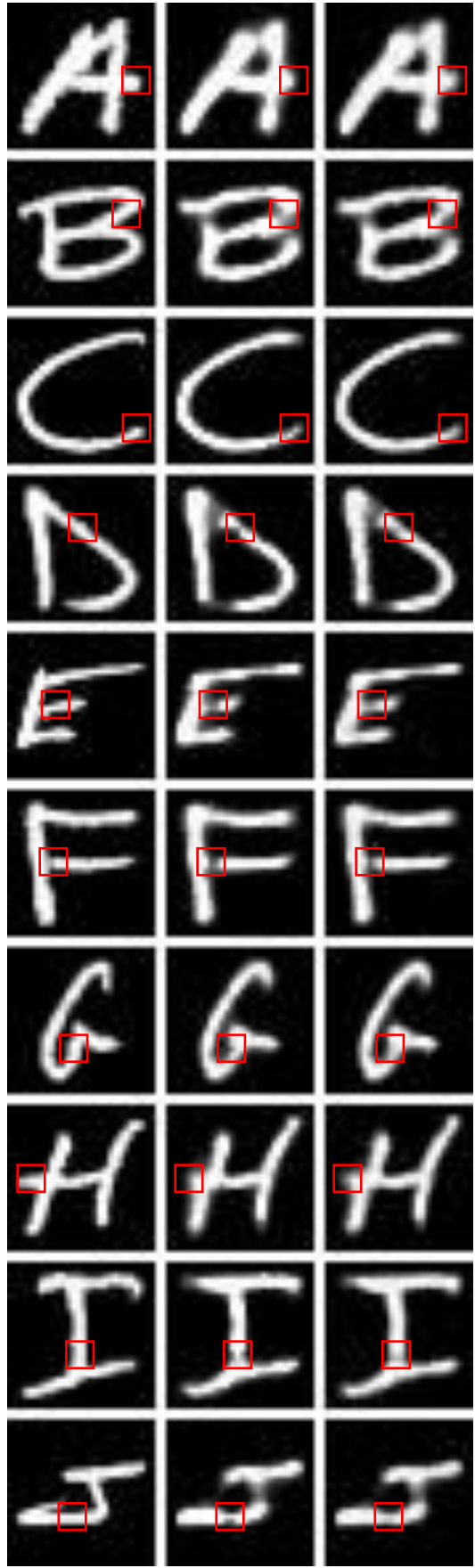

**Figure 20.** Marked reconstruction images on EMNIST dataset (big letters).

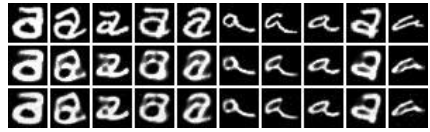

(1)

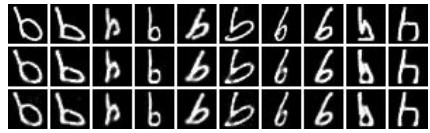

(2)

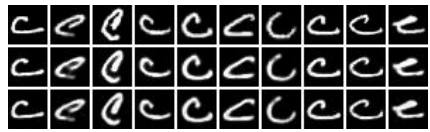

(3)

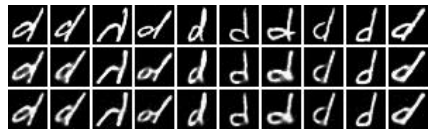

(4)

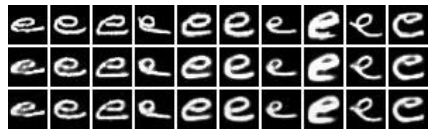

(5)

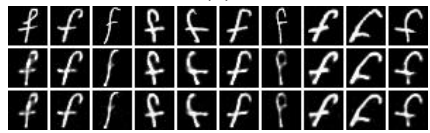

(6)

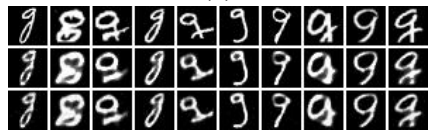

(7)

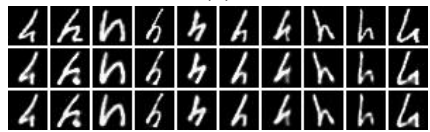

(8)

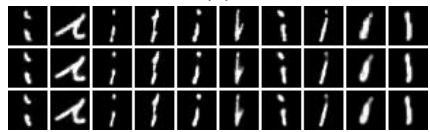

(9)

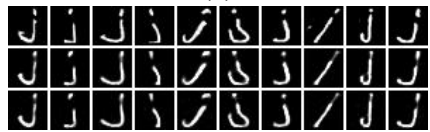

(10)

**Figure 21.** Reconstruction images on EMNIST dataset (small letters).

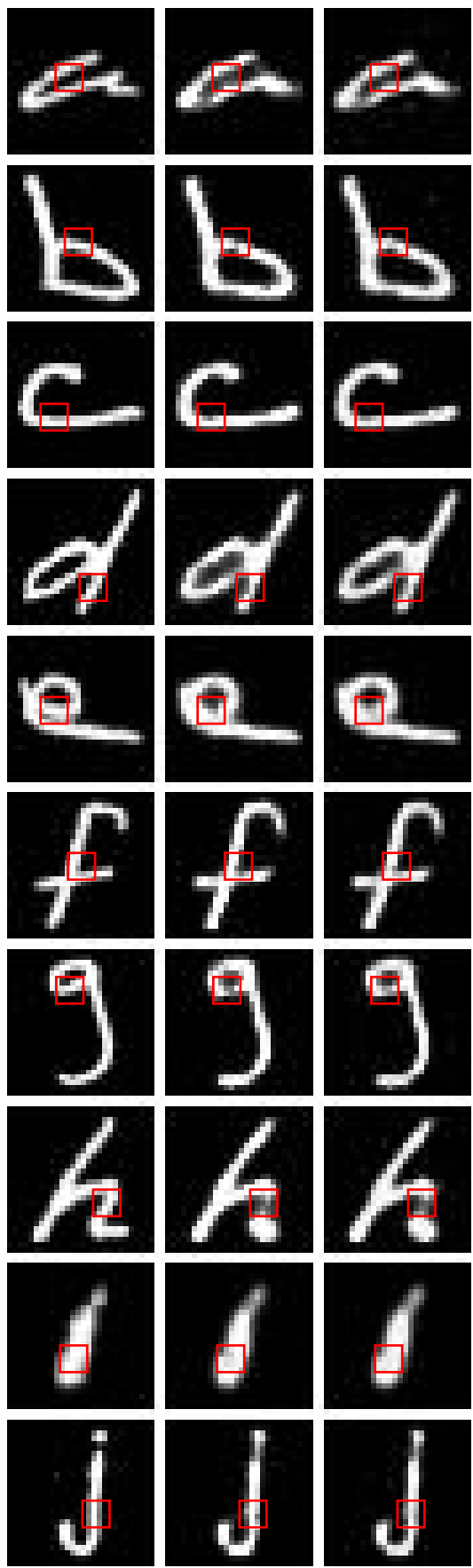

**Figure 22.** Marked reconstruction images on EMNIST dataset (small letters).

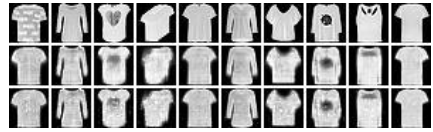
(1)
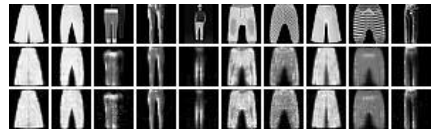
(2)
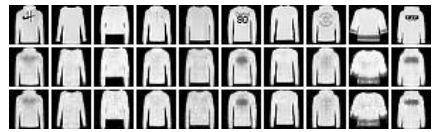
(3)
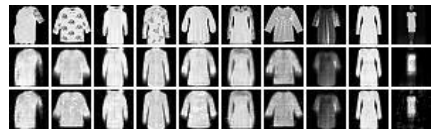
(4)
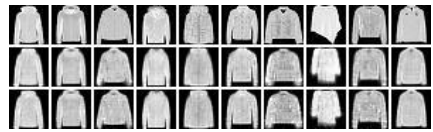
(5)
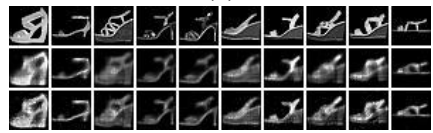
(6)
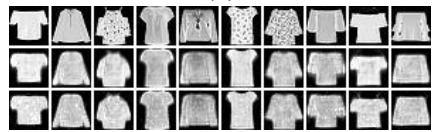
(7)
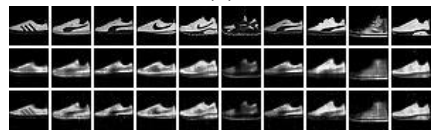
(8)
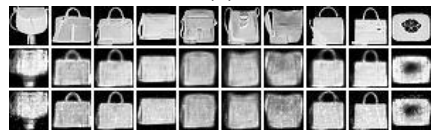
(9)
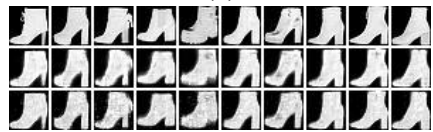
(10)

**Figure 23.** Reconstruction images on FMNIST dataset.

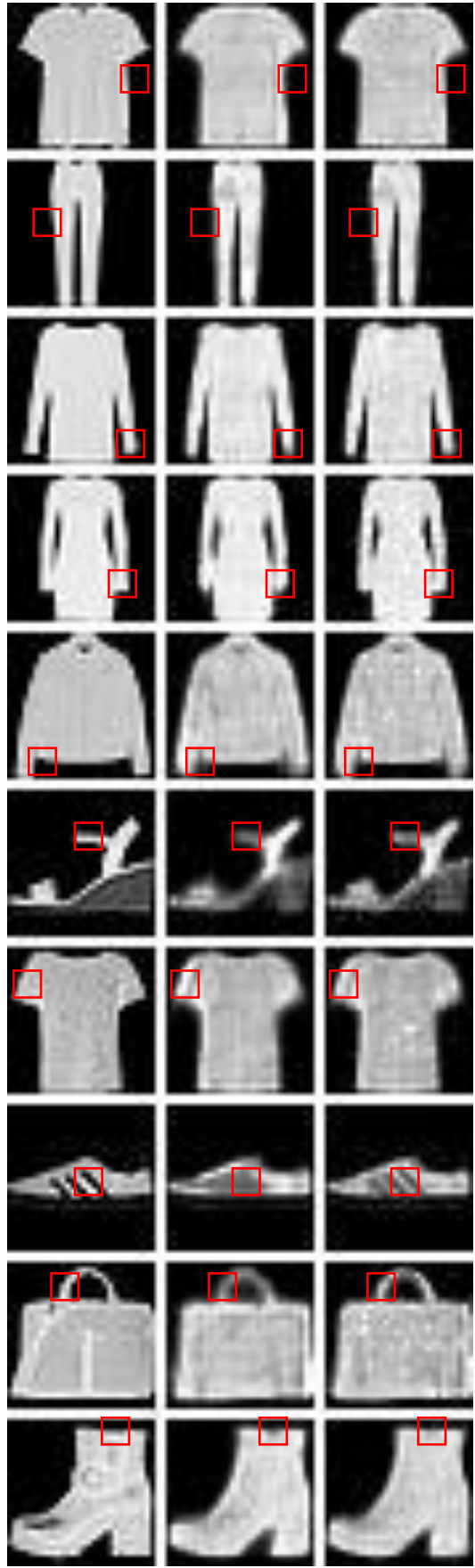

**Figure 24.** Marked reconstruction images on FMNIST dataset.

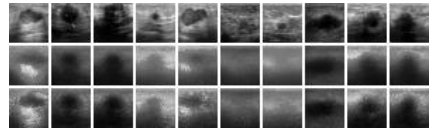

(1)

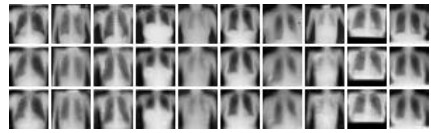

(2)

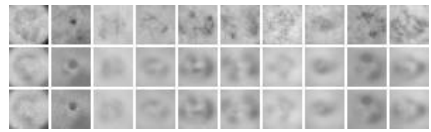

(3)

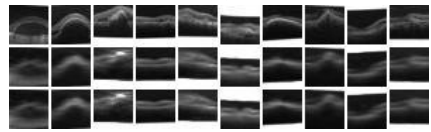

(4)

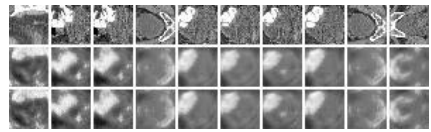

(5)

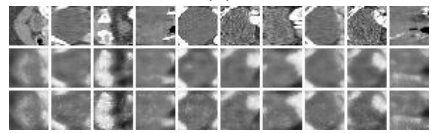

(6)

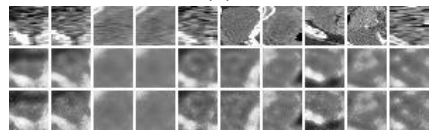

(7)

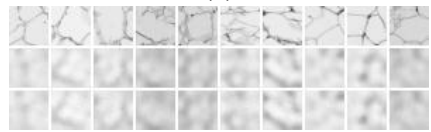

(8)

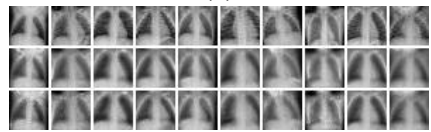

(9)

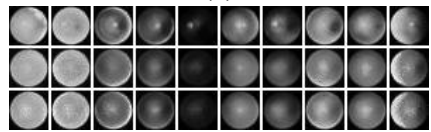

(10)

**Figure 25.** Reconstruction images on MMNIST dataset.

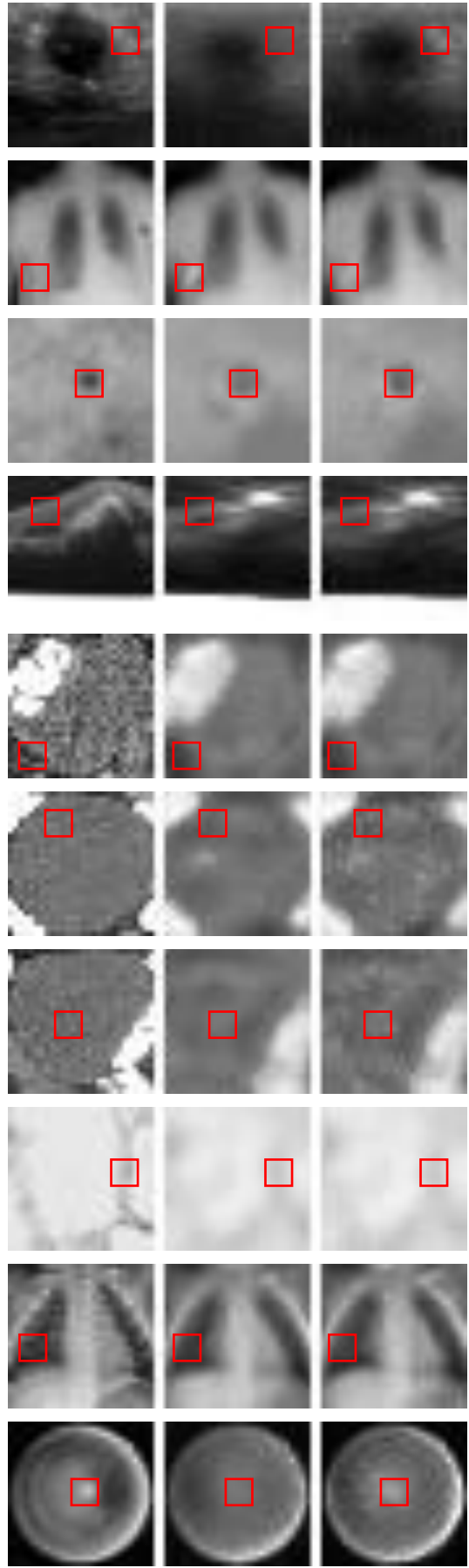

**Figure 26.** Marked reconstruction images on MMNIST dataset.

In our future work, experiments on datasets with large resolution images and colorful images will be conducted. The experiments on other advanced autoencoders, such as VAE and WAE, will be explored. The convolutional layer or transformer layer will be introduced in the proposed algorithms. The constrains on high-dimensional reconstruction data, such as sparse and low-rank priors, will be utilized to advance the reconstruction performance of autoencoders. The generalized autoencoders-based data compression and signal compressed sensing will also be probed. The autoencoders-based lossless reconstruction will further be studied.

## 6. Patents

The patent with application number CN202110934815.7 and publication number CN113642709A results from the work reported in this manuscript.

**Supplementary Materials:** The following supporting information can be downloaded at: github.com/xxx/s1, Figure S1: title; Table S1: title; Video S1: title.

**Author Contributions:** Conceptualization, Honggui Li and Maria Trocan; methodology, Honggui Li and Dimitri Galayko; writing, Honggui Li; supervision, Mohamad Sawan.

**Funding:** This research received no external funding.

**Institutional Review Board Statement:** Not applicable.

**Informed Consent Statement:** Not applicable.

**Data Availability Statement:** Not applicable.

**Acknowledgments:** The authors would very much like to thank Yui Chun Leung for the collection of MATLAB implementations of Generative Adversarial Networks (GANs) on the GitHub website (https://github.com/zcemycl/Matlab-GAN). We take advantage of the AAE codes and halve the dimension of AAE latent space..

**Conflicts of Interest:** The authors declare no conflict of interest.

## Biographies

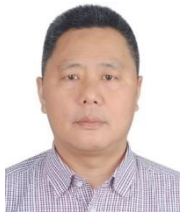

**Honggui Li:** received a B.S. degree in electronic science and technology from Yangzhou University, and received a Ph.D. degree in mechatronic engineering from Nanjing University of Science and Technology. He is a senior member of the Chinese Institute of Electronics. He is a visiting scholar and a post-doctoral fellow in Institut Supérieur d'Électronique de Paris for one year. He is an associate professor of electronic science and technology and a postgraduate supervisor of electronic science and technology at Yangzhou University. He is a reviewer for some international journals and conferences. He is the author of over 30 refereed journal and conference articles. His current research interests include machine learning, deep learning, computer vision, and embedded computing.

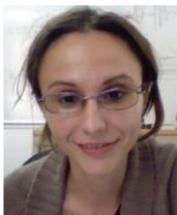

**Maria Trocan:** received her M.Eng. in Electrical Engineering and Computer Science from Politehnica University of Bucharest, the Ph.D. in Signal and Image Processing from Telecom ParisTech and the Habilitation to Lead Researches (HDR) from Pierre & Marie Curie University (Paris 6). She has joined Joost - Netherlands, where she worked as research engineer involved in the design and development of video transcoding systems. She is firstly Associate Professor, then Professor at Institut Superieur d'Electronique de Paris (ISEP). She is Associate Editor for Springer Journal on Signal, Image and Video Processing and Guest Editor for several journals (Analog Integrated Circuits and Signal Processing, IEEE Communications Magazine etc.). She is an active member of IEEE France and served as counselor for the ISEP IEEE Student Branch, IEEE France Vice-President responsible of Student Activities and IEEE Circuits and Systems Board of Governors member, as Young Professionals representative. Her current research interests focus on image and video analysis & compression, sparse signal representations, machine learning and fuzzy inference.

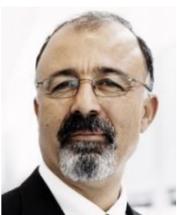

**Mohamad Sawan:** (Fellow, IEEE) received the Ph.D. degree in electrical engineering from the University of Sherbrooke, Sherbrooke, QC, Canada, in 1990. He was a Chair Professor awarded with the Canada Research Chair in Smart Medical Devices (2001–2015), and was leading the Microsystems Strategic Alliance of Quebec - ReSMiQ (1999–2018). He is a Professor of Microelectronics and Biomedical Engineering, in leave of absence from Polytechnique Montréal, Canada. He joined Westlake University, Hangzhou, China, in January 2019, where he is a Chair Professor, Founder, and the Director of the Center for Biomedical Research And INnovation (CenBRAIN). He has published more than 800 peer-reviewed articles, two books, ten book chapters, and 12 patents. He founded and chaired the IEEE-Solid State Circuits Society Montreal Chapter (1999–2018) and founded the Polystim Neurotech Laboratory, Polytechnique Montréal (1994–present), including two major research infrastructures intended to build advanced Medical devices. He is the Founder of the International IEEE-NEWCAS Conference, and the Co-Founder of the International IEEE-BioCAS, ICECS, and LSC conferences. He is a Fellow of the Canadian Academy of Engineering, and a Fellow of the Engineering Institutes of Canada. He is also the "Officer" of the National Order of Quebec. He has served as a member of the Board of Governors (2014–2018). He is the Vice-President Publications (2019–present) of the IEEE CAS Society.

He received several awards, among them the Queen Elizabeth II Golden Jubilee Medal, the Barbara Turnbull 2003 Award for spinal-cord research, the Bombardier and Jacques-Rousseau Awards for academic achievements, the Shanghai International Collaboration Award, and the medal of merit from the President of Lebanon for his outstanding contributions. He was hosted in Montreal as the General Chair, the 2016 IEEE International Symposium on Circuits and Systems (ISCAS), and hosted as the General Chair of the 2020 IEEE International Medicine, Biology and Engineering Conference (EMBC). Dr. Sawan was the Deputy Editor-in-Chief of the IEEE TRANSACTIONS ON CIRCUITS AND SYSTEMS-II: EXPRESS BRIEFS (2010–2013); the Co-Founder, an Associate Editor, and the Editor-in-Chief of the IEEE TRANSACTIONS ON BIOMEDICAL CIRCUITS AND SYSTEMS; an Associate Editor of the IEEE TRANSACTIONS ON BIOMEDICALS ENGINEERING; and the International Journal of Circuit Theory and Applications.

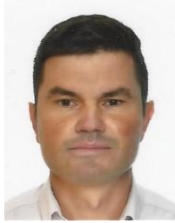

**Dimitri Galayko:** received the bachelor's degree from Odessa State Polytechnic University in Ukraine, the master's degree from the Institute of Applied Sciences of Lyon in France, and the Ph.D. degree from University Lille in France. He made his Ph.D. thesis in the Institute of Microelectronics and Nanotechnologies. His Ph.D. dissertation was on the design of micro-electromechanical silicon filters and resonators for radio-communications. He is an Associate Professor with the LIP6 research laboratory of Sorbonne University in France. His research interests include study, modelling, and design of nonlinear integrated circuits for sensor interface and for mixed-signal applications. His research interests also include machine learning and fuzzy computing.